\documentclass[sigconf,authorversion,nonacm]{acmart}

\usepackage{balance}
\usepackage[utf8]{inputenc}
\def\BibTeX{{\rm B\kern-.05em{\sc i\kern-.025em b}\kern-.08emT\kern-.1667em\lower.7ex\hbox{E}\kern-.125emX}}

\usepackage{longtable}
\usepackage{tabularx}
\usepackage{afterpage}
\usepackage{placeins}
\usepackage{rotating}
\usepackage{array}
\usepackage{todonotes}
\usepackage{xspace}
\usepackage[linesnumbered,ruled,vlined]{algorithm2e}
\usepackage{adjustbox}
\usepackage[english]{babel}
\usepackage{soul}
\usepackage{flushend}
\usepackage{listings}
\usepackage{amsfonts}
\usepackage{amsmath}
\usepackage{graphicx}
\usepackage{subcaption}
\usepackage{multirow}
\usepackage{etoolbox}\AtBeginEnvironment{algorithmic}{\small} 
\theoremstyle{definition}

\usepackage{mathtools}
\DeclarePairedDelimiter\ceil{\lceil}{\rceil}
\DeclareMathOperator*{\argmax}{arg\,max}

\usepackage{makecell}

\newcommand{\sysname}[0]{Morphence}

\makeatletter
\patchcmd{\@makecaption}
  {\scshape}
  {}
  {}
  {}

\newcommand*{\da@rightarrow}{\mathchar"0\hexnumber@\symAMSa 4B }
\newcommand*{\da@leftarrow}{\mathchar"0\hexnumber@\symAMSa 4C }
\newcommand*{\xdashrightarrow}[2][]{%
  \mathrel{%
    \mathpalette{\da@xarrow{#1}{#2}{}\da@rightarrow{\,}{}}{}%
  }%
}
\newcommand{\xdashleftarrow}[2][]{%
  \mathrel{%
    \mathpalette{\da@xarrow{#1}{#2}\da@leftarrow{}{}{\,}}{}%
  }%
}
\newcommand*{\da@xarrow}[7]{%
  \sbox0{$\ifx#7\scriptstyle\scriptscriptstyle\else\scriptstyle\fi#5#1#6\m@th$}%
  \sbox2{$\ifx#7\scriptstyle\scriptscriptstyle\else\scriptstyle\fi#5#2#6\m@th$}%
  \sbox4{$#7\dabar@\m@th$}%
  \dimen@=\wd0 %
  \ifdim\wd2 >\dimen@
    \dimen@=\wd2 %
  \fi
  \count@=2 %
  \def\da@bars{\dabar@\dabar@}%
  \@whiledim\count@\wd4<\dimen@\do{%
    \advance\count@\@ne
    \expandafter\def\expandafter\da@bars\expandafter{%
      \da@bars
      \dabar@ 
    }%
  }%
  \mathrel{#3}%
  \mathrel{%
    \mathop{\da@bars}\limits
    \ifx\\#1\\%
    \else
      _{\copy0}%
    \fi
    \ifx\\#2\\%
    \else
      ^{\copy2}%
    \fi
  }%
  \mathrel{#4}%
}
\makeatother

\newcolumntype{M}[1]{>{\centering\arraybackslash}p{#1}}

\usepackage{enumitem}
\setlist[itemize,1]{leftmargin=1.5\parindent, itemsep=0ex, topsep=0.5ex, 
}
\setlist[enumerate,1]{leftmargin=2\parindent, itemsep=0ex, topsep=0.5ex, 
}

\setcopyright{none} 
\begin{document}

\title{Morphence: Moving Target Defense Against Adversarial Examples}

\begin{abstract}
Robustness to adversarial examples of machine learning models remains an open topic of research. Attacks often succeed by repeatedly probing a {\em fixed target} model with adversarial examples purposely crafted to fool it. 
In this paper, we introduce \sysname{}, an approach that shifts the defense landscape by making a model a {\em moving target} against adversarial examples. By regularly moving the decision function of a model, \sysname{} makes it significantly challenging for repeated or correlated attacks to succeed. \sysname{} deploys a pool of models generated from a base model in a manner that introduces sufficient randomness when it responds to prediction queries. To ensure repeated or correlated attacks fail, the deployed pool of models automatically expires after a query budget is reached and the model pool is seamlessly replaced by a new model pool generated in advance. 
We evaluate \sysname{} on two benchmark image classification datasets (MNIST and CIFAR10) against five reference attacks (2 white-box and 3 black-box). In all cases, \sysname{} consistently outperforms the thus-far effective defense, adversarial training, even in the face of strong  white-box attacks, while preserving accuracy on clean data and reducing attack transferability.
\end{abstract}

\author{Abderrahmen Amich}
\affiliation{%
 \institution{University of Michigan, Dearborn}
   \country{}
   }
 \email{aamich@umich.edu}

\author{Birhanu Eshete}
\affiliation{%
  \institution{University of Michigan, Dearborn}
  \country{}}
  \email{birhanu@umich.edu}

\settopmatter{printfolios=true}
\maketitle

\section{Introduction}\label{sec: intro}
Machine learning (ML) continues to propel a broad range of applications in image classification~\cite{ImageNet}, voice recognition~\cite{DL-Speech2012}, precision medicine~\cite{DeepCC2019}, malware/intrusion detection~\cite{malconv18}, autonomous vehicles~\cite{DL-autnonmous17}, and so much more. ML models are, however, vulnerable to {\em adversarial examples} ---minimally perturbed legitimate inputs that fool models to make incorrect predictions~\cite{FGSM,Biggio-ECML13}. Given an input $x$ (e.g., an image) correctly classified by a model $f$, an adversary performs a small perturbation $\delta$ and obtains $x' = x+\delta$ that is indistinguishable from $x$ to a human analyst, yet the model misclassifies $x'$. Adversarial examples pose realistic threats on domains such as self-driving cars and malware detection for the consequences of incorrect predictions are highly likely to cause real harm~\cite{AV-Physical-Attack17,MalConvEvade18,MalGAN17}.

To defend against adversarial examples, previous work took multiple directions each with its pros and cons. Early attempts~\cite{Early-Defense14,Early-Defense15} to harden ML models provided only marginal robustness improvements against adversarial examples. Heuristic defenses based on {\em defensive distillation}~\cite{distillation}, {\em data transformation}~\cite{Compression17,Compression18,Augmentation17,Cropping17,Rand15,Rand18}, and {\em gradient masking}~\cite{Thermo-Encode18,PixelDefend18} were subsequently broken \cite{Carlini-Breaking17,Carlini-BreakingUsenix17,Gradient-Masking18,CW}.

While {\em adversarial training}~\cite{FGSM,EnsembelAdvTrain18} remains effective against known classes of attacks, robustness comes at a cost of accuracy penalty on clean data. Similarly, data transformation-based defenses also penalize prediction accuracy on legitimate inputs. \textit{Certified defenses}~\cite{lecuyer2019certified,RandomSmoothing19,Certified-AdditiveNoise19} provide formal robustness guarantee, but are limited to a class of attacks constrained to LP-norms \cite{lecuyer2019certified,wong2018provable}.

As pointed out by Goodfellow~\cite{goodfellow2019research}, a shared limitation of all prior defense techniques is the {\em static and fixed target} nature of the deployed ML model. We argue that, although defended by methods such as adversarial training, the very fact that a ML model is a {\em fixed target} that continuously responds to prediction queries makes it {\em a prime target for repeated/correlated adversarial attacks}. As a result, given enough time, an adversary will have the advantage to repeatedly query the prediction API and build enough knowledge about the ML model and eventually fool it. Once the adversary launches a successful attack, it will be always effective since the model is not moving from its compromised ``location''. 

In this paper, we introduce \sysname{}, an approach that makes a ML model a moving target in the face of adversarial example attacks. By regularly moving the decision function of a model, \sysname{} makes it significantly challenging for an adversary to successfully fool the model through adversarial examples. Particularly, \sysname{} significantly reduces the effectiveness of once successful and repeated attacks and attacks that succeed after patiently probing a fixed target model through correlated sequence of attack queries. \sysname{} deploys a pool of $n$ models generated from a base model in a manner that introduces sufficient randomness when it selects the most suitable model to respond to prediction queries. The selection of the {\em most suitable model} is governed by a scheduling strategy that relies on the prediction confidence of each model on a given query input. To ensure repeated or correlated attacks fail, the deployed pool of $n$ models automatically expires after a query budget is reached. The model pool is then seamlessly replaced by a new pool of $n$ models generated and queued in advance. In order to be practical, \sysname{}'s moving target defense (MTD) strategy needs to address the following challenges:
\begin{itemize}
    \item {\bf Challenge-1:} Significantly improving \sysname{}'s robustness to adversarial examples across white-box and black-box attacks.
\item {\bf Challenge-2:} Maintaining accuracy on clean data as close to that of the base model as possible.
\item {\bf Challenge-3:} Significantly increasing diversity among models in the pool to reduce adversarial example transferability among them.
\end{itemize}

\sysname{} addresses {\bf Challenge-1} by enhancing the MTD aspect through: larger model pool size, a model selection scheduler, and dynamic pool renewal (Sections \ref{subsec:model-gen} and \ref{subsec:renewal}). {\bf Challenge-2} is addressed by re-training each generated model to regain accuracy loss caused by perturbations (Section \ref{subsec:model-gen}: step-2). \sysname{} addresses {\bf Challenge-3} by making the individual models distant enough via distinct transformed training data used to re-train each model (Section \ref{subsec:model-gen}: step-2). Training a subset of the generated models on distinct adversarial data is an additional robustness boost to address {\bf Challenge-1 and 3} (Section \ref{subsec:model-gen}: step-3).

We note that while \sysname{} is not the first MTD-based approach to tackle adversarial examples, it advances the state-of-the-art in MTD-based prior work ~\cite{fMTD19,MTDeep19,EI-MTD} by making the MTD core more resilient against adversarial examples, significantly reducing accuracy loss on clean data, and limiting adversarial example transferability among models that power the MTD strategy.  In Section \ref{subsec: related}, we precisely position \sysname{} against closely related work.

We evaluate \sysname{} on two benchmark image classification datasets (MNIST and CIFAR10) using five attacks: two white-box attacks (FGSM~\cite{FGSM}, C\&W~\cite{CW}) and three black-box attacks (SPSA~\cite{uesato2018adversarial}, Copycat~\cite{CopyCat18}+FGSM, Copycat~\cite{CopyCat18}+C\&W). We compare \sysname{}'s robustness with adversarial training defense of a fixed model. We then conduct detailed evaluations on the impact of the MTD strategy in defending previously successful repeated attacks. Additionally, through extensive experiments, we shed light on each component of \sysname{} and its impact towards improving the robustness results and reduce the transferability rate across models. Overall, our evaluations suggest that \sysname{} advances the state-of-the-art in robustness against adversarial examples, even in the face of strong white-box attacks such as C\&W~\cite{CW}, while maintaining accuracy on clean data and reducing attack transferability. In summary, this paper builds on the successes of MTD in network and software security~\cite{MTD-book} and makes the following contributions:
\begin{itemize}
\item \sysname{} improves the current state-of-the-art on robustness against adversarial examples, both white-box and black-box.

\item \sysname{} automatically thwarts repeated attacks that leverage previously successful attacks and correlated attacks performed through dependent consecutive queries.

\item \sysname{} outperforms static adversarial training in terms of robustness against adversarial examples, while maintaining accuracy on clean data and reducing attack transferability within a model pool.

\item \sysname{} code is available as free and open source software at: {\color{blue} \url{https://github.com/um-dsp/Morphence}}.
\end{itemize}

The rest of this paper is organized as follows. In Section \ref{sec: bground}, we introduce adversarial examples and put \sysname{} in the context of related work. In Section \ref{sec: approach}, we present technical details of \sysname{}. Section \ref{sec: eval} presents extensive evaluations of \sysname{}. Finally, Section \ref{sec: concl} concludes the paper.
\section{Background and Related Work}\label{sec: bground}
To streamline our discussion in subsequent sections, we first briefly introduce adversarial examples and put related work in context.

\subsection{Background: Adversarial Examples}
\textbf{Machine Learning Basics.}
In supervised ML, for a set of labeled training samples $X_{train} = (X_i,y_i): i\le n$, where $X_i$ is a training example and $y_i$ is the corresponding label, the objective of training a ML model $\theta$ is to minimize the expected loss over all $(X_i,y_i): J(\theta) = \frac{1}{n}\sum_{i=1}^{n}l(\theta,X_i,y_i)$. In models such as logistic regression and deep neural networks, the loss minimization problem is typically solved using stochastic gradient descent (SGD) via iterative update of $\theta$ as:
  \begin{equation}\label{eq:gradient-descent}
     \theta = \theta - \epsilon\cdot \Delta_{\theta}\sum_{i=1}^{n} l(\theta,X_i,y_i)
  \end{equation}
 where $\Delta_{\theta}$ is the gradient of the loss with respect to the weights $\theta$; $X$ is a randomly selected set (e.g., {\em mini-batches}) of training examples drawn from $X_{train}$; and $\epsilon$ is the {\em learning rate} which controls the magnitude of change on $\theta$.

 Let $X$ be a $d$-dimensional feature space and $Y$ be a $k$-dimensional output space, with underlying probability distribution $Pr(X,Y)$, where $X$ and $Y$ are random variables for the feature vectors and the classes (labels) of data, respectively. The objective of testing a ML model is to perform the mapping  $f_{\theta} : X \rightarrow Y$. The output of $f_{\theta}$ is a $k$-dimensional vector and each dimension represents the probability of input belonging to the corresponding class. 
 
\textbf{Adversarial Examples.}
Given a ML model with a decision function $f:X \rightarrow Y$ that maps an input sample $x \in X$ to a true class label $y_{true} \in Y$, $x'$ = $x  + \delta$ is called an {\em adversarial example} with an {\em adversarial perturbation} $\delta$ if: 
\begin{equation}
    f(x') = y' \ne y_{true}, ||\delta|| < \epsilon
\end{equation}
where $||.||$ is a distance metric (e.g., one of the $L_{p}$ norms)  and $\epsilon$ is the maximum allowable perturbation that results in misclassification while preserving semantic integrity of $x$. Semantic integrity is domain and/or task specific. For instance, in image classification, visual imperceptibility of $x'$ from $x$ is desired while in malware detection $x$ and $x'$ need to satisfy certain functional equivalence (e.g., if $x$ was a malware pre-perturbation, $x'$ is expected to exhibit maliciousness post-perturbation as well).
In {\em untargeted} evasion, the goal is to make the model misclassify a sample to any different class. When {\em targeted}, the goal is to make the model to misclassify a sample to a specific target class.

Adversarial examples can be crafted in {\em white-box} or {\em black-box} setting. Most gradient-based attacks~\cite{FGSM,BIM,PGSM,CW} are white-box because the adversary typically has access to model details, which allow to query the model directly to decide how to increase the model’s loss function. Gradient-based attacks assume that the adversary has access to the gradient function of the model. The goal is to find the perturbation vector $\delta^\star \in \mathbb{R}^d$ that maximizes the loss function $J(\theta, x, y_{target})$ of the model $f$, where $\theta$ are the parameters (i.e., weights) of the model $f$. In recent years, several white-box attacks have been proposed, especially for image classification tasks. Some of the most notable ones are: Fast Gradient Sign Method (FGSM)~\cite{FGSM}, Basic Iterative Method (BIM)~\cite{BIM}, Projected Gradient Descent (PGD) method~\cite{PGSM}, and Carlini \& Wagner (C\&W) method~\cite{CW}. Black-box attack techniques (e.g., MIM~\cite{MIM}, HSJA~\cite{HSJA20}, SPSA~\cite{uesato2018adversarial}) begin with initial perturbation $\delta_{0}$, and probe $f$ on a series of perturbations $f (x + \delta_{i})$, to craft $x'$ such that $f$ misclassifies it to a label different from its original. 
 
Prior work has also shown the feasibility to approximate a black-box model via model extraction attacks~\cite{Model-Stealing,orekondy2018knockoff,CopyCat18, AdvApprox20}. The extracted model is then leveraged as a white-box proxy to craft adversarial examples that evade the target black-box model\cite{Practical-black-box16}. In this sense, model extraction plays the role of a steppingstone for adversarial example attacks. In this paper, we use two white-box attacks (FGSM and C\&W) and three black-box attacks (SPSA, Copycat+FGSM, and Copycat+C\&W). We provide details of these attacks in the Appendix (Section \ref{subsec:attacks}).

 \subsection{Related Work}\label{subsec: related}
We review related work along three lines: best-effort heuristic defenses, certified defenses, and moving target defenses (the subject of this paper).

\textbf{Best-Effort Heuristic Defenses.}
In this class of defenses, several defense approaches have been proposed, most of which were subsequently broken \cite{Carlini-Breaking17,Carlini-BreakingUsenix17,Gradient-Masking18}. Many of the early attempts~\cite{Early-Defense14,Early-Defense15} to harden ML models (especially neural networks) provided only marginal robustness improvements against adversarial examples. In the following, we highlight this line of defenses.

{\em Defensive distillation} by Papernot et al.~\cite{distillation} is a strategy to distill knowledge from neural network as soft labels by smoothing the {\em softmax} layer of the original training data. It then uses the soft labels to train a second neural network that, by way of hidden knowledge transfer, would behave like the first neural network. The key insight is that by training to match the first network, one will hopefully avoid over-fitting against any of the training data. Defensive distillation was later broken by Carlini and Wagner~\cite{CW}.

{\em Adversarial training}~\cite{FGSM}: While effective against the class of adversarial examples a model is trained against, it fails to catch adversarial examples not used during adversarial training. Moreover, robustness comes at a cost of accuracy penalty on clean inputs.

{\em Data transformation approaches} such as compression~\cite{Compression17,Compression18}, augmentation~\cite{Augmentation17}, cropping~\cite{Cropping17}, and randomization~\cite{Rand15,Rand18} have also been proposed to thwart adversarial examples. While these lines of defenses are effective against attacks constructed without the knowledge of the transformation methods, all it takes an attacker to bypass such countermeasures is to employ novel data transformation methods. Like adversarial training, these approaches also succeed at a cost of accuracy loss on clean data.

{\em Gradient masking approaches} (e.g., \cite{Thermo-Encode18,PixelDefend18}) are based on obscuring the gradient from a white-box adversary. However, Athalye et al.~\cite{Gradient-Masking18} later broke multiple gradient masking-based defenses.

\textbf{Certifiably Robust Defenses.}
Best-effort defenses remain vulnerable because adaptive adversaries work around their defensive heuristics. To  obtain a theoretically justified guarantee of robustness, certified defenses have been recently proposed~\cite{wong2018provable,raghunathan2020certified,lecuyer2019certified,RandomSmoothing19}. The key idea is to provide provable/certified robustness of ML models under attack.

Lecuyer et al.\cite{lecuyer2019certified} prove robustness lower bound with differential privacy using Laplacian and Gaussian noise after the first layer of a neural network. Li et al.~\cite{Certified-AdditiveNoise19} build on~\cite{lecuyer2019certified} to prove certifiable lower bound using Renyi divergence. To stabilize the effect of Gaussian noise, Cohen et al.~\cite{RandomSmoothing19} propose randomized smoothing with Gaussian noise to guarantee adversarial robustness in $l_2$ norm. They do so by turning any classifier that performs well under Gaussian noise into a new classifier that is certifiably robust to adversarial perturbations under $l_2$ norm. 

\textbf{Moving Target Defenses.}
Network and software security has leveraged numerous flavors of MTD including randomization of service ports and address space layout randomization~\cite{MTD-Survey18,MTD-book,Morpheus19}. Recent work has explored MTD for defending adversarial examples~\cite{fMTD19,MTDeep19,EI-MTD}.

Song et al.~\cite{fMTD19} proposed a fMTD where they create fork-models via independent perturbations of the base model and retrain them. Fork-models are updated periodically whenever the system is in an idle state. The input is sent to all models and the prediction label is decided by majority vote. 

Sengupta et al. \cite{MTDeep19} proposed  MTDeep, which uses different DNN architectures (e.g., CNN, HRNN, MLP) in a manner that reduces transferability between model architectures using a measure called {\em differential immunity}. Through Bayesian Stackelberg game, MTDeep chooses a model to classify an input. Despite diverse model architectures, MTDeep suffers from a small model pool size.

Qian et al.~\cite{EI-MTD} propose EI-MTD, a defense that leverages the Bayesian Stackelberg game for dynamic scheduling of student models to serve prediction queries on resource-constrained edge devices. The student models are generated via differential distillation from an accurate teacher model that resides on the cloud. 

\textbf{\sysname{} vs. Prior MTD-Based Work}. Next, we specifically position \sysname{} with respect to MTD-based related work~\cite{fMTD19,MTDeep19,EI-MTD}.\\
\textit{Compared to EI-MTD~\cite{EI-MTD}}, \sysname{} avoids inheritance of adversarial training limitations by adversarially training a subset of student models instead of the base model. Unlike EI-MTD that results in lower accuracy on clean data after adversarial training, \sysname{}'s accuracy on clean data after adversarial training is much better since the accuracy penalty is not inherited by student models. Instead of adding regularization term during training, \sysname{} uses distinct transformed training data to retrain student models and preserve base model accuracy. Instead of the Bayesian Stackelberg game, \sysname{} uses the most confident model for prediction. 

\textit{With respect to MTDeep~\cite{MTDeep19}}, \sysname{} expands the pool size regardless of the heterogeneity of individual models and uses average transferability rate to estimate attack transferability. On scheduling strategy, instead of Bayesian Stackelberg game \sysname{} uses the most confident model. 

\textit{Unlike fMTD~\cite{fMTD19}}, \sysname{} goes beyond retraining perturbed fork models and adversarially trains a subset of the model pool to harden the whole pool against adversarial example attacks. In addition, instead of majority vote, \sysname{} picks the most confident model for prediction. For pool renewal, instead of waiting when the system is idle, \sysname{} takes a rather safer and transparent approach and renews an expired pool seamlessly on-the-fly.
\section{Approach}\label{sec: approach}

\begin{figure*}[t!]
    
    \centering
    \scalebox{.88}{
    \includegraphics[width=\textwidth]{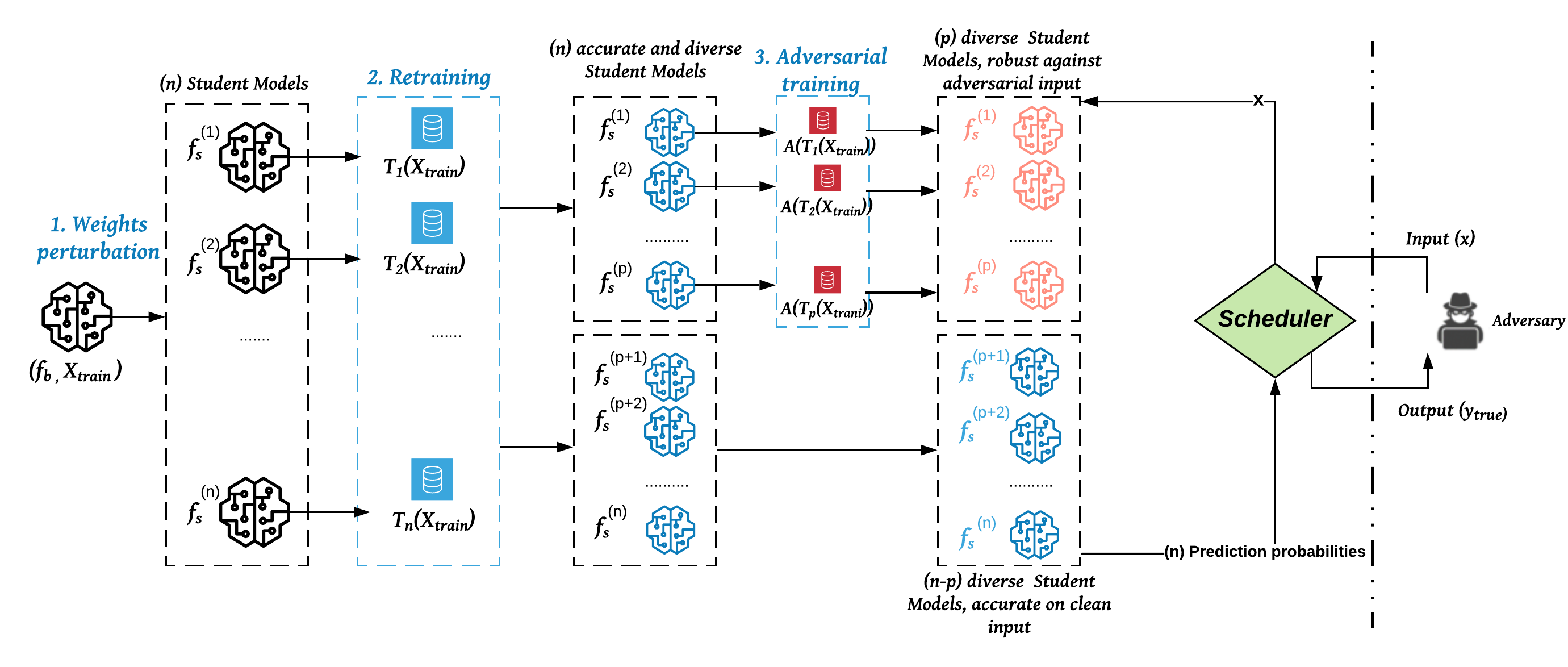}}
    \caption{Illustration for the generation of a pool of $n$ models in \sysname{}.}
    \label{fig:rmtd}

\end{figure*}

We first describe \sysname{} at a high level in Section \ref{subsec:approach-overview} and then dive deeper into details in Sections \ref{subsec:model-gen} and \ref{subsec:renewal}. We use Algorithm \ref{alg:model_gen} and Algorithm  \ref{alg:model_retrain} to illustrate technical details of \sysname{}.

\subsection{Overview} \label{subsec:approach-overview}
The key intuition in \sysname{} is making a model a moving target in the face of adversarial example attacks. To do so, \sysname{} deploys a pool of $n$ models instead of a single target model. We call each model in pool a {\em student model}. The model pool is generated from a base model in a manner that introduces sufficient randomness when \sysname{} responds to prediction queries, without sacrificing accuracy of the base model on clean inputs. In response to each prediction query, \sysname{} selects the {\em most suitable model} to predict the label for the input.  The selection of the {\em most suitable model} is governed by a scheduling strategy ({\em Scheduler} in Figure \ref{fig:rmtd}) that relies on the prediction confidence of each model on a given input. The deployed pool of $n$ models automatically expires after a query budget $Q_{max}$ is exhausted. An expired model pool is seamlessly replaced by a new pool of $n$ models generated and queued in advance. As a result, repeated/correlated attacks fail due to the moving target nature of the model.
In the following, we will use Figure \ref{fig:rmtd} to highlight the core components of \sysname{} focusing on \textit{model pool generation}, \textit{model pool renewal}, and the main motivations behind these components of \sysname{}.

\textbf{Model Pool Generation.}
The model pool generation method is composed of three main steps. Each step aims to tackle one or more of challenges 1--3. As shown in Figure \ref{fig:rmtd}, a highly accurate base model $f_b$ initially trained on a training set $X_{train}$ is the foundation from which $n$ student models that make up the \sysname{} student model pool are generated. Each student model is initially obtained by slightly perturbing the parameters of $f_b$ (Step-1, \textit{weights perturbation}, in Figure \ref{fig:rmtd}). By introducing different and random perturbations to model parameters, we {\em move} the base model's decision function into $n$ different {\em locations} in the prediction space.

Given the randomness of the weights perturbation, Step-1 is likely to result in inaccurate student models compared to the base model. Simply using such inaccurate models as part of the pool penalizes prediction accuracy on clean inputs. As a result, we introduce Step-2 to address {\bf Challenge-2} by \textit{retraining} the student models to boost their respective accuracy and bring it close enough to the base model's accuracy. However, since the $n$ models are generated from the same base model, their transferability rate of adversarial examples is expected to be high. We take measures to reduce transferability of evasion attacks across the student models by using distinct training sets for each student model in Step-2 (the {\em retraining} step). For this purpose, we harness data transformation techniques (i.e., image transformations) to produce $n$ distinct training sets. In Section \ref{eval:components}, we empirically explore to what extent this measure reduces the transferability rate and addresses {\bf Challenge-3}. Finally, a subset $p \leq n$  of student models is adversarially trained (Step-3 in Figure \ref{fig:rmtd}) as a reinforcement to the MTD aspect of \sysname{} against adversarial queries, which addresses {\bf Challenge-1}. In Section \ref{subsec:model-gen}, we explain the motivations and details of each step.

\textbf{Scheduling and Pool Renewal.}
Instead of randomly selecting a model from $n$ deployed models or take the majority vote of the $n$ models~\cite{fMTD19}, our scheduling strategy is such that for a given input query the prediction of the {\em most confident model} is returned by \sysname{}. The key intuition behind selecting the most confident model is the sufficient diversity among the $n$ models: where a subset of the models is pre-hardened with adversarial training (hence perform much better on adversarial inputs) and the remaining models are trained to perform more confidently on legitimate inputs. \\
The model pool is automatically renewed after a defender-set query upper-bound $Q_{max}$ is reached. The choice of $Q_{max}$ requires careful consideration of the model pool size ($n$) and the time it takes to generate a new model pool while \sysname{} is serving prediction queries on an active model pool. In Section \ref{subsec:renewal}, we describe the details of the model pool renewal with respect to $Q_{max}$.

\subsection{Student Model Pool Generation}\label{subsec:model-gen}
Using Algorithms \ref{alg:model_gen} and \ref{alg:model_retrain}, we now describe the details of steps 1--3 with respect to challenges 1--3.

\textbf{Step-1: Model Weights Perturbation.}
Shown in Step-1 of Figure \ref{fig:rmtd}, the first step to transform the fixed model $f_b$ into a moving target is the generation of multiple instances of $f_b$. To effectively serve the MTD purpose, the generated instances of $f_b$ need to fulfill two conditions.  First, they need to be {\em sufficiently diverse} to reduce adversarial example transferability among themselves. Second, they need to preserve the accuracy of the base model.

By applying $n$ different small perturbations on the model weights $\theta_b$ of $f_b$, we generate $n$ variations of $f_b$ as $f_s =\{f_s^{(1)}, f_s^{(2)}, ..., f_s^{(n)} \}$, and we call each  $f_s^{(i)}$ a {\em student model}. The $n$ perturbations should be sufficient to produce $n$ diverse models that are different from the base model $f_b$. More precisely, higher perturbations lead to a larger distance between the initial base model $f_b$ and a student model $f_s^{(i)}$, which additionally contribute to greater movement of the decision function of the base model. However, the $n$ perturbations are constrained by the need to preserve the prediction accuracy of the base model $f_b$ for each student model $f_s^{(i)}$. As shown in Algorithm \ref{alg:model_gen} (line 2), we perturb the parameters $\theta_b$ by adding noise sampled from the \textit{Laplace} distribution.

Laplace distribution is defined as $\frac{1}{2\lambda}\exp(-\frac{|\theta_b-\mu|}{\lambda})$ \cite{Laplace}. The center of the post-perturbation weights distribution is the original weights $\theta_b$. We fix the mean value of the added Laplace noise as $\mu = 0$ (line 2). The perturbation bound defined by the Laplace distribution is $exp(\lambda)$, which is a function of the noise scale $\lambda$, also called the exponential decay. Our choice of the Laplace mechanism is motivated by the way the exponential function scales multiplicatively, which simplifies the computation of the multiplicative bound (i.e., $exp(\lambda)$).

However, there is no exact method to find the maximum noise scale $\lambda_{max}$>0 that guarantees acceptable accuracy of a generated student model. Thus, we approximate $\lambda_{max}$ empirically with respect to the candidate student model by incrementally using higher values of $\lambda>0$ until we obtain a maximum value $\lambda_{max}$ that results in a student model with unacceptable accuracy. Additionally, we explore the impact of increasing $\lambda$ on the overall performance of the prediction framework and the transferability rate of evasion attacks across the $n$ models (Section \ref{eval:components}). We note that the randomness of Laplace noise allows the generation of $n$ different student models using the same noise scale $\lambda$. Furthermore, even in case of a complete disclosure of the fabric of our approach, it is still difficult for an adversary to reproduce the exact pool of $n$ models to use for adversarial example generation, due to the random aspect of the model generation approach.

\begin{algorithm}[t!]
\small
\KwResult{$f_s$ : student model}
\SetAlgoLined
\textbf{Input:}\break
$f_b$: base model;\break
$X_{train}$ : training set;\break
$X_{test}$ : testing set;\break
$acc_b \gets Accuracy(f_b, X_{test})$;\break
$T_i$ : data transformation function;\break
$\lambda>0$ : noise scale;\break
$\epsilon > 0$ : used to detect the convergence of model training;\break
$max\_acc\_loss$ : allowed margin of accuracy loss between $f_b$ and $f_s$;\break
$adv\_train$ : boolean variable that indicates whether to train student model on adversarial data;\break
$\Lambda$ : mixture of evasion attacks to use for adversarial training when $adv\_train = TRUE$;\break
\textbf{Step-1:}\tcp{model weights perturbation.}\break
$f_s \gets f_b + Lap(0,\lambda)$;\break
\tcp{$Lap(\mu,\lambda)$ returns an array of noise samples drawn from Laplace distribution $\frac{1}{2\lambda}\exp(-\frac{|\theta_b-\mu|}{\lambda})$}

\textbf{Step-2:}\tcp{retraining on transformed data.}\break
$f_s \gets$ retrain($f_s$, $T_i(X_{train})$, $X_{test}$, $\epsilon$, $Adv=FALSE$);\break
$acc_s \gets $Accuracy($f_s$, $X_{test}$);\break
\While{$acc_b - acc_s$ > $max\_acc\_loss$}
{ repeat Step-1 and Step-2 with smaller $\lambda$;}
\textbf{Step-3:}\tcp{retraining on adversarial data.}\break
\If{$adv\_train$}{$f_s \gets $retrain($f_s$, $\Lambda(T_i(X_{train}))$, $X_{test}$, $\epsilon$, $Adv=TRUE$);\break
\tcp{check accuracy loss on clean test set.}\break
$acc_s \gets$ Accuracy($f_s$, $X_{test}$);\break
\While{$acc_b - acc_s > max\_acc\_loss$}
{$f_s \gets$ retrain($f_s$, $T_i(X_{train})$, $X_{test}$, $\epsilon$, $Adv=FALSE$);\break}}

\caption{Generation of a Student Model.}
\label{alg:model_gen}
\end{algorithm}

\textbf{Step-2: Retraining on Transformed Data.}
Minor distortions of the parameters of $f_b$ have the potential to drastically decrease the prediction accuracy of the resultant student model. Consequently, Step-1 is likely to produce student models that are less accurate than the base model. An accuracy recovery measure is necessary to ensure that each student model has acceptable accuracy close enough to the base model. To that end, we retrain the $n$ newly created student models (line 3).

We recall that the diversity of the $n$ models is crucial for \sysname{}'s MTD core

such that adversarial examples are less transferable across models ({\bf Challenge-3}). In this regard, retraining all student models on $X_{train}$ used for the base model $f_b$ results in models that are too similar to the decision function of the base model $f_b$. It is, therefore, reasonable to use a distinct training set for each student model. To tackle data scarcity, we harness data augmentation techniques to perform $n$ distinct transformations $\{T_1,...,T_n\}$ on the original training set $X_{train}$ (e.g., translation, rotation, etc). The translation distance or the rotation degree are randomly chosen with respect to the validity constraint of the transformed set $T_i(X_{train})$. A transformed sample is valid only if it is still recognized by its original label. In our case, for each studied dataset (i.e., MNIST, CIFAR10), we use benchmark transformations proposed and validated by previous work \cite{Tian2018DetectingAE}. We additionally double-check the validity of the transformed data by verifying whether each sample is correctly recognized by the base model.

\begin{algorithm}[t!]
\small
\KwResult{$f_s^{(i)}$ : student model}
\SetKwFunction{Fretrain}{Retrain}
\SetAlgoLined
\SetKwProg{Fn}{Def}{:}{}
  \Fn{\Fretrain{$f_s^{(i)}$, $X_{retrain}$, $X_{test}$, $\epsilon$, $Adv$}}{
  \tcp{For adversarial training we use adversarial test examples for validation.}\break
  \If{$Adv=TRUE$}{$X_{test}\gets\Lambda(X_{test})$}
  $acc_{tmp} \gets$ Accuracy($f_s^{(i)}$, $X_{test}$);\break
  $epochs \gets 0$;\break
  \While{TRUE}{
  $f_s^{(i)}$.train($X_{retrain}$, $epoch=1$);\break
  $acc \gets$ Accuracy($f_s^{(i)}$, $X_{test}$);\break
  \tcp{check training convergence.}\break
  \If{$epochs$ mod($5$) $=0$}{
    \eIf{$|acc - acc_{tmp}|<\epsilon$}{break;\break}
    {$acc_{tmp}\gets acc$;}
    }\break
    $epochs \gets epochs + 1$;\break
    }
   }

\caption{Student model retraining function.}
\label{alg:model_retrain}
\end{algorithm}
Algorithm \ref{alg:model_retrain} illustrates the student model retraining function called \texttt{Retrain($f_s$,$X_{retrain}$,$X_{test}$,$\epsilon$,$Adv$)}. It takes as inputs a student model to retrain $f_s$, a retraining set $X_{retrain} = T_i(X_{train})$, a testing set $X_{test}$, a small positive infinitesimal quantity $\epsilon \rightarrow 0$ used for training convergence detection, and a boolean flag $Adv$. As indicated on line 15 of Algorithm \ref{alg:model_gen}, $Adv$ is $FALSE$ since the retraining data does not include adversarial examples. The algorithm regularly checks for training convergence after a number of (e.g., 5) epochs. The retraining convergence is met when the current accuracy improvement is lower than $\epsilon$ (lines 7--13 in Algorithm \ref{alg:model_gen}).

The validity of the selected value of the noise scale $\lambda$ used in Step-1 is decided by the outcome of Step-2 in regaining the prediction accuracy of a student model. More precisely, if retraining the student model (i.e., Step-2) does not improve the accuracy, then the optimisation algorithm that minimizes the model's loss function is stuck in a local minimum due to the significant distortion brought by weight perturbations performed in Step-1. In this case, we repeat Step-1 using a lower $\lambda$, followed by Step-2. The loop breaks when the retraining succeeds to regain the student model's accuracy, which indicates that the selected $\lambda$ is within the maximum bound $\lambda < \lambda_{max}$ (Algorithm \ref{alg:model_gen}, lines 4-6). For more control over the accuracy of deployed models, we define a hyperparameter $max\_acc\_loss$, that is configurable by the defender. It represents the maximum prediction accuracy loss tolerated by the defender.
\\
\\

\textbf{Step-3: Adversarial Training a Subset of $p$ Student Models.} 

To motivate the need for Step-3, we first assess the current design using just Step-1 and Step-2 with respect to challenges 1--3.

Suppose \sysname{} is deployed based only on Step-1 and Step-2, and for each input it picks the most confident student model and returns its prediction. On clean inputs, the MTD strategies introduced in Step-1 (via model weights perturbation) and Step-2 (via retraining on transformed training data) make \sysname{}'s prediction API a moving target with nearly no loss on prediction accuracy. On adversarial inputs, an input that evades student model $f^{(i)}_s$ is less likely to also evade another student model $f^{(j)}_s$ because of the significant reduction of transferability between student model predictions as a result of Step-2. However, due to the exclusive usage of clean inputs in Step-1 and Step-2, an adversarial example may still fool a student model of \sysname{} on first attempt. We note that the success rate of a repeated evasion attack is low because the randomness introduced in Step-1 disarms the adversary of a stable fixed target model that returns the same prediction for repeated queries on a given adversarial input. To significantly reduce the success of one-step evasion attacks via adversarial inputs, we introduce selective adversarial training to reinforce the MTD strategy built through Step-1 and Step-2. More precisely, we perform adversarial training on a subset $p \leq n$ models from the $n$ student models obtained after Step-2 (lines 7--13 in Algorithm \ref{alg:model_gen}). It is noteworthy that while our choice of adversarial training is based on the current state of the art, conceptually, a defender is free to use a different (possibly better) method than adversarial training. 

\textbf{Why adversarial training on $p \leq n$ student models?} 
There are three intuitive alternatives to integrate adversarial training to the MTD strategy: $(a)$ adversarial training of the base model $f_b$ before Step-1; $(b)$ adversarial training  of all $n$ student models after Step-2; or $(c)$ adversarial training of a subset of student models after Step-2. As noted by prior work \cite{EI-MTD}, $(a)$ is bound to result in an inherited robustness for each student model, which costs less execution time compared to adversarial training of $n$ student models. However, in this case, the inherent limitation of adversarial training, i.e., accuracy reduction on clean inputs, is also inherited by the $n$ student models. Alternative $(b)$ suffers from similar drawbacks. In particular, by adversarially training all $n$ models, while making individual student models resilient against adversarial examples, we risk sacrificing the overall prediction accuracy on clean data. Considering the drawbacks of $(a)$ and $(b)$, we pursue $(c)$. In particular, we select $p\leq n$ student models for adversarial training (lines 7--13 in Algorithm \ref{alg:model_gen}). Consequently, the remaining $n-p$ models remain as accurate as the base model on clean data in addition to being diverse enough to reduce transferability among them. 

\textbf{Adversarial training approach.} 
We now explain the details of adversarially training a student model with respect to Algorithm \ref{alg:model_gen}.
Once again, the \texttt{Retrain} function illustrated in Algorithm \ref{alg:model_retrain} is invoked using different inputs (line 11). For instance, $X_{retrain}=\Lambda(T_i(X_{train})$, which indicates that $f_s$ is trained on adversarial examples, is generated by performing a mixture of evasion attacks $\Lambda$ on the transformed training set $T_i(X_{train})$ specified for student model $f_{s}^{(i)}$. To reduce accuracy decline on clean data, we shuffle the training set with additional clean samples from $T_i(X_{train})$. Furthermore, we use more than one evasion attack with different perturbation bounds $||\delta||<\xi$ for adversarial samples generation to boost the robustness of the student model against different attacks (e.g., C\&W~\cite{CW}, gradient-based~\cite{FGSM}, etc). Like Step-2, the training convergence is reached if the improvement of the model's accuracy on adversarial examples (i.e., the robustness) recorded periodically, i.e., after a number (e.g., 5) of epochs, becomes infinitesimal (lines 7--13 in Algorithm \ref{alg:model_retrain}).

\textbf{How to choose the values of $p$ and $n$?} The values $p$ and $n$ are hyper-parameters of \sysname{} and are chosen by the defender. Ideally, larger value of $n$ favors the defender by creating a wider space of movement for a model's decision function (thus creating more uncertainty for repeated or correlated attacks). However, in practice $n$ is conditioned by the computational resources available to the defender. Therefore, here we choose not to impose any specific values of $n$.  We, however, recall that $Q_{max}$ is proportional with the time needed to generate a pool of $n$ models. Therefore, it is plausible that $n$ need not be too large to the extent that it leads to a long extension of the expiration time of the pool of $n$ models due to a longer period $T_n$ needed to generate $n$ models that causes a large value of $Q_{max}$. Regarding the number of adversarially-trained models ($p$), there is no exact method to select an optimal value. Thus, we empirically examine all possible values of $p$ between $0$ and $n$ to explore the performance change of \sysname{} on clean inputs and adversarial examples. Additionally, we explore the impact of the value $p$ on the transferability rate across the $n$ student models (results are discussed in Section \ref{eval:components}).

\textbf{Scheduling Strategy.}
In the following, what we mean by {\em scheduling strategy} is the act of selecting the model that returns the label for a given input to \sysname{}'s prediction API. There are multiple alternatives to reason about the scheduling strategy. Randomly selecting a model or taking the majority vote of all student models are both intuitive avenues. However, random selection does not guarantee effective model selection and majority vote does not consider the potential inaccuracy of adversarially trained models on clean queries.

\textit{Most Confident Model.} In \sysname{}, we rather adopt a strategy that relies on the confidence of each student model. More precisely, given an input $x$, \sysname{} first queries each student model and returns the prediction of the most confident model. Given a query $x$, the scheduling strategy is formally defined as: $\argmax \{f_s^{(1)}(x),...,f_s^{(n)}(x)\}$.

In section \ref{eval:robust}, we investigate the effectiveness of the scheduling strategy in increasing the overall robustness of \sysname{} against adversarial examples. 
\vspace{-1em}
\subsection{Model Pool Renewal} \label{subsec:renewal}
Given that $n$ is a finite number, with enough time, an adversary can ultimately build knowledge about the prediction framework through multiple queries using different types of inputs. For instance, if the adversary correctly guesses model pool size $n$, it is possible to map which model is being selected for each query by closely monitoring the prediction patterns of multiple examples. Once compromised, the whole framework becomes a sitting target since its movement is limited by the finite number $n$.

An intuitive measure to avoid such exposure is abstaining from responding to a ``suspicious" user \cite{goodfellow2019research}. However, given the difficulty to precisely decide whether a user is suspect based solely on the track of its queries, this approach has the potential to result in a high abstention rate, which unnecessarily leads to denial of service for legitimate users. We, therefore, propose a relatively expensive yet effective method to ensure the continuous mobility of the target model without sacrificing the quality of service. In particular, we actively update the pool of $n$ models when a query budget upper bound $Q_{max}$ is reached. To maintain the quality of service in terms of query response time, the update needs to be seamless. To enable seamless model pool update, we ensure that a buffer of batches of $n$ student models is continuously generated and maintained on stand-by mode for subsequent deployments.

The choice of $Q_{max}$ is crucial to increase the mobility of the target model while satisfying the condition:
\textit{"the buffer of pools of models is never empty at a time of model batch renewal"}.\\ 
Suppose at time $t$ the buffer contains $K_t$ pools of models. A new pool is removed from the buffer after every period of $Q_{max}$ queries and a clean-slate pool that responds to prediction queries is activated. Thus, the buffer is exhausted after $K_t.Q_{max}$ number of queries. Furthermore, we suppose that the response time to a single query $q$ is $T_q$ and the generation of a pool of $n$ models lasts a period of $T_n$. Then, the above condition is formally expressed as:
$\begin{array}{rrclcl}
\displaystyle
K_t.Q_{max}.T_q > T_n \textrm{, s.t.} & K_t > 0.
\end{array}$
The inequality implies that the expected time to exhaust the whole buffer of pools must be always greater than the duration of creating a new pool of $n$ models. Additionally, it shows that $Q_{max}$ is variable with respect to the time $t$ and the number of models in one pool $n$. Ideally, $Q_{max}$ should be as low as possible to increase the mobility rate of target model. Thus, the optimal solution is $\ceil {\frac{T_n}{K_t.T_q} }$.

\section{Evaluation}\label{sec: eval}
We now present \sysname{}'s evaluation.
Section \ref{subsec:setup} presents our experimental setup. Section \ref{eval:robust} compares \sysname{} with the best empirical defense to date: adversarial training. Section \ref{mtd_impact} examines the impact of dynamic scheduling and model pool renewal. Finally, Section \ref{eval:components} sheds light on the impact of each component of \sysname{} on robustness and attack transferability.

\subsection{Experimental Setup}
\label{subsec:setup}
 In this section, we describe the experimental setup adopted for \sysname{} evaluation. Particularly, we explain the studied datasets, evasion attacks, base models, and  baseline defenses. Additionally, we report the employed evaluation metrics and the hyper parameters configuration of \sysname{} framework.
 
\textbf{Datasets:}
We use two benchmark datasets: MNIST~\cite{MNIST} and CIFAR10~\cite{cifar}. We use $5$K test samples of each dataset to perform $5$K queries on \sysname{} to evaluate its performance on clean inputs and its robustness against adversarial examples. 

MNIST~\cite{MNIST}: The MNIST dataset contains 70, 000 grey-scale images of hand-written digits in 10 class labels (0--9). Each image is normalized to a dimension of 28x28 pixels.

CIFAR10~\cite{cifar}: The CIFAR10 dataset contains 60, 000 color images of animals and vehicles in 10 class labels (0--9). The classes include labels such as bird, dog, cat,
airplane, ship, and truck. Each image is 32x32 pixels in dimension.

\textbf{Attacks:}
We use two \textit{white-box} attacks (C\&W~\cite{CW} and FGSM~\cite{FGSM}) and three \textit{black-box} attacks (first we use an iterative black-box attack SPSA \cite{uesato2018adversarial}, then we use a model extraction attack Copycat \cite{CopyCat18} to perform C\&W and FGSM on the extracted substitute model). Details of all the attacks appear in the Appendix (Section \ref{subsec:attacks}). For C\&W and FGSM, we assume the adversary has white-box access to $f_b$. For SPSA, Copycat+ C\&W, and Copycat+ FGSM, the adversary has oracle access to \sysname{}. We carefully chose each attack to assess our contribution claims. For instance, C\&W, being one of the most effective white-box attacks, is strongly suggested as a benchmark for ML robustness evaluation~\cite{carlini2019evaluating}. FGSM is fast and salable on large datasets, and it generalizes across models ~\cite{kurakin2017adversarial}. In addition, its high transferability rate across models makes it suitable to investigate the effectiveness of \sysname{}'s different components to reduce attack transferability across student models. To explore \sysname{}'s robustness against query-based black-box attacks, we employ SPSA since it performs multiple correlated queries before crafting adversarial examples.
For all attacks we use a perturbation bound $||\xi||_\infty < 0.3$.

For adversarial training, we adopt PGDM \cite{PGSM} and C\&W \cite{CW} to create a mixed set of adversarial training data (i.e., $\Lambda(X_{train})$ in Algorithm \ref{alg:model_gen}). On CIFAR10, we exclude C\&W to speed up adversarial training. We generate adversarial training samples using a perturbation bound $||\xi||_\infty < 0.5$. Hyper-parameter configuration details of each attack are provided in the Appendix (Section \ref{hyper}).

\textbf{Base Models:}
For MNIST, we train a state-of-the-art 6-layer CNN model that reaches a test accuracy of $99.72\%$ on a test set of 5K. We call this model ``MNIST-CNN". As for CIFAR10, we use the same CNN architecture proposed in the Copycat paper \cite{CopyCat18}. It reaches an accuracy of $83.63\%$ on a test set of 5K. We call this model ``CIFAR10-CNN". More details about both models are provided in the Appendix (Section \ref{subsec:base-models}).

\textbf{Baseline Defenses:}
For both datasets, we compare \sysname{} with an \textit{undefended fixed} model and an \textit{adversarially-trained fixed} model. Adversarial training has been considered as one of the most effective and practical defenses against evasion attacks. However, it sacrifices the model's accuracy on clean data. In Table \ref{tab:evasion_res}, we show to what extent \sysname{} reduces accuracy loss caused by adversarial training while improving robustness against adversarial examples.

\textbf{Hyper-parameters:}
As explained in Section \ref{sec: approach}, setting up \sysname{} for defense requires tuning parameters. Results in Table \ref{tab:evasion_res} and discussions in Section \ref{eval:robust} are based on pool size of $n=10$. We use $5$ pools of student models generated before \sysname{} begins to respond to queries. Since $X_{test}$=5K, the maximum number of queries to be received by \sysname{} is 5K. Thus, for the sake of local evaluations, $Q_{max}$ is fixed as 1K. Hence, each 1K samples of $X_{test}$=5K are classified by a distinct pool of models.

In Section \ref{eval:components}, for each dataset, we conduct multiple experiments using different values of $p<n$ to empirically find out the values of $p<n$ that result in better performance. Furthermore, we explore different configurations of the \textit{noise scale} $\lambda>0$ to detect the threshold $\lambda_{max}$ that bounds the weights perturbation step performed on the base model $f_b$ to generate the student models. We recall that using $\lambda>\lambda_{max}$ leads to much greater distortion of model weights that in turn leads to unacceptably low accuracy student model. Accordingly, for MNIST, we fix $\lambda=0.1$, $p=5$ and for CIFAR10, we use  $\lambda=0.05$ and we explore the different results of $p=8$ and $p=9$. More discussion about the impact of $\lambda>0$ and $p<n$ on \sysname{}'s performance are provided in Section \ref{eval:components}.

\textbf{Evaluation Metrics:}
We now define the main evaluation metrics used in our experiments.

\textit{\underline{Accuracy}.}
We compute \sysname{}'s accuracy on adversarial data using the different attacks described earlier and compare it with  baseline fixed models. 

\textit{\underline{Average Transferability Rate}.}
We compute the average transferability rate across all $n$ models to evaluate the effectiveness of the data transformation measures in \sysname{} at reducing attack transferability. To compute the transferability rate from model $f^{(i)}_s$ to another model $f^{(j)}_s$, we calculate the rate of adversarial examples that succeeded on $f^{(i)}_s$ that also succeed on $f^{(j)}_s$. Across all student models, the \textit{Average Transferability Rate} is computed as:

$\displaystyle 
\frac{1}{n (n-1)}\sum_{i=1}^{n} \sum_{\substack{j=1 \\ j\ne i}}^{n} \frac{N_{adv}(f^{(i)}_s \rightarrow f^{(j)}_s)}{N_{adv}(f^{(i)}_s)},$\\

where $N_{adv}(f^{(i)}_s) = \{x' \in X'_{test}; f^{(i)}_s(x') \ne y_{true}\}$ is the number of adversarial examples that fooled $f^{(i)}_s$ and $N_{adv}(f^{(i)}_s \rightarrow f^{(j)}_s) = \{x' \in N_{adv}(f^{(i)}_s); f^{(j)}_s(x') \ne y_{true}\}$ is the number of adversarial examples that fooled $f^{(i)}_s$ that still fool $f^{(j)}_s$.

\begin{table*}[t!]
\centering
  \scalebox{0.83}{
   \begin{tabular}{|c!{\color{black}\vrule}c!{\color{black}\vrule}c!{\color{black}\vrule}!{\color{black}\vrule}!{\color{black}\vrule}c!{\color{black}\vrule}c!{\color{black}\vrule}c!{\color{black}\vrule}c!{\color{black}\vrule}}
       \hline
       
& \multicolumn{3}{c!{\color{black}\vrule}}{\textbf{MNIST-CNN Accuracy}} & \multicolumn{3}{c!{\color{black}\vrule}}{\textbf{CIFAR10-CNN Accuracy}}\\
\hline
\textbf{Attack}& \textbf{Undefended} & \textbf{\thead{Adversarially Trained} } & \textbf{\sysname} & \textbf{Undefended} &  \textbf{\thead{Adversarially Trained}} & \thead{\textbf{\sysname} ($p=8$, $p=9$)}\\ 
       \hline
       No Attack & 99.72\%& 97.17\% &\textbf{99.04\%}& 83.63\%& 75.37\% &\textbf{84.64\%}, \textbf{82.65\%} \\ 
       \hline\hline
       FGSM~\cite{FGSM} & 9.98\%&  42.38\% &\textbf{71.43\%}&19.98\%&36.62\%&\textbf{36.44\%}, \textbf{38.78\%}\\ 
       \hline
       C\&W~\cite{CW} & 0.0\%& 0.0\% &\textbf{97.75\%}&1.25\%&1.34\%&\textbf{44.50\%}, \textbf{40.91\%}\\ 
       \hline
       SPSA~\cite{uesato2018adversarial} &29.04\% & 59.43\%&\textbf{97.77\%}&38.96\%&59.08\%&\textbf{60.85\%}, \textbf{62.83\%} \\ 
       \hline
       \hline
        Copycat~\cite{CopyCat18} + FGSM~\cite{FGSM} &40.33\% & 72.75\%&\textbf{74.06\%}&16.95\%&39.26\%&\,\textbf{39.75\%,  40.81\%} \\ 
        \hline
        Copycat~\cite{CopyCat18} + C\&W~\cite{CW} &97.04\% & 98.63\%&\textbf{98.82\%}&51.36\%&54.72&\,\textbf{58.12\%, 55.03\%}\\ 
       \hline
   \end{tabular}}
 
 \vspace*{1em}
\caption{\sysname{} prediction accuracy under white-box and black-box attacks, compared to an undefended and adversarially trained fixed model.}
\label{tab:evasion_res}
\end{table*}
\vspace{-1em}
\subsection{Robustness Against Evasion Attacks}
\label{eval:robust}
Based on results summarized in Table \ref{tab:evasion_res}, we now evaluate the robustness of \sysname{} compared to an \textit{undefended fixed model} and \textit{adversarially trained} fixed model across the 5 reference attacks.

\textbf{Robustness in a nutshell:} Across all attacks and threat models, \sysname{} is more robust than \textit{adversarial training} for both datasets. In particular, \sysname{} significantly outperforms adversarial training by an average of $\approx 55\%$ on MNIST across all three studied attacks (i.e., FGSM, C\&W and SPSA). On CIFAR10, for FGSM and SPSA, \sysname{} achieves comparable robustness to adversarial training if $p=8$, while it shows an increase of $\approx 2\%$ when $p=9$. With regards to C\&W, it shows a drastic improvement in robustness compared to the baseline models (i.e., $\approx 41\%$).

\textbf{Accuracy loss on clean data:} Table \ref{tab:evasion_res} indicates that, unlike adversarial training on a fixed model, \sysname{} does not sacrifice accuracy to improve robustness. For instance, while adversarial training drops the accuracy of the undefended MNIST-CNN by $\approx 3\%$, \sysname{} maintains it in par with its original value ($>99\%$). Similar results are observed on CIFAR10. In particular, even after using $9$ adversarially-trained models ($p=9$) of $n=10$ student models for each pool, the accuracy loss is $\leq1\%$, while it reaches $8.26\%$ for an adversarially-trained CIFAR10-CNN. This result is due to the effectiveness of the adopted scheduling strategy to assign clean queries mostly to student models that are confident on clean data. 
Furthermore, \sysname{} improved the original accuracy for lower values of $p$. For instance, Table \ref{tab:evasion_res} shows an improvement of $1\%$ in the accuracy on clean data compared to the undefended baseline model, on CIFAR10 for $p=8$.
In conclusion, our findings indicate that \textit{\sysname{} is much more robust compared to adversarial training on fixed model, while maintaining (if not improving) the original accuracy of the undefended base model}.

\textbf{Robustness against C\&W:} Inline with the state-of-the-art, Table \ref{tab:evasion_res} shows that C\&W is highly effective on the baseline fixed models. For both datasets, even after adversarial training, C\&W attack is able to maintain its high attack accuracy for both datasets ($100\%$ on MNIST and $\approx 99\%$ on CIFAR10). However, it significantly fails achieve the same attack accuracy on \sysname{}. For instance, \sysname{} increases the robustness against C\&W data by $\approx 97\%$ for MNIST and $\approx40\%$ for CIFAR10 compared to adversarial training on fixed model. This significant improvement in robustness is brought by the moving target aspect included in \sysname{}. More precisely, given the low transferability rate of C\&W examples across different models, C\&W queries that are easily effective on the base model are not highly transferable to the student models, hence less effective on \sysname{}. More discussion that reinforces this insight is presented in Section \ref{eval:components}.

\textbf{Robustness against FGSM:} Although known to be a less effective attack than C\&W on the fixed models, FGSM performs better on \sysname. For instance, \sysname's accuracy on FGSM data is $26.32\%$ less for MNIST and $\approx 3\%$ less for CIFAR10, compared to C\&W. These findings are explained by the high transferability rate of FGSM examples across student models and the base model. However, despite the transferability issue \sysname{} still improves robustness on FGSM data. More discussion about the transferability effect on \sysname{} are provided in Section \ref{eval:components}.

\textbf{Robustness against SPSA (an iterative query-based attack):} For a more realistic threat model, we turn to a case where the adversary has a black-box access to \sysname{}. Through multiple interactions with \sysname{}'s prediction API, the attacker tries different perturbations of an input example to reach the evasion goal. Accordingly, SPSA performs multiple queries using different variations of the same input to reduce the SPSA loss function. Table \ref{tab:evasion_res} shows that \sysname{} is more robust on SPSA than the two baseline fixed models. Thanks to its dynamic characteristic, \sysname{} is a moving target. Hence, it can derail the iterative query-based optimization performed by SPSA. More analysis into the impact of the dynamic aspect is discussed in Section \ref{mtd_impact}.

\textbf{Robustness against model extraction-based attacks:} The last 2 rows in Table \ref{tab:evasion_res} show that \sysname{} is slightly robust than adversarial training on C\&W and FGSM attacks generated using a Copycat surrogate model. Clearly, adversarial examples generated on Copycat models in a black-box setting are much less effective than the ones generated in white-box setting, which is inline with the state-of-the-art. We also notice that FGSM is more effective on Copycat models than C\&W, which is explained by the relatively higher transferability rate of FGSM examples from the Copycat model to the target model, compared to the low transferability of C\&W examples.

\begin{figure*}[t!]
    
    \centering   
    \includegraphics[width=.97\textwidth]{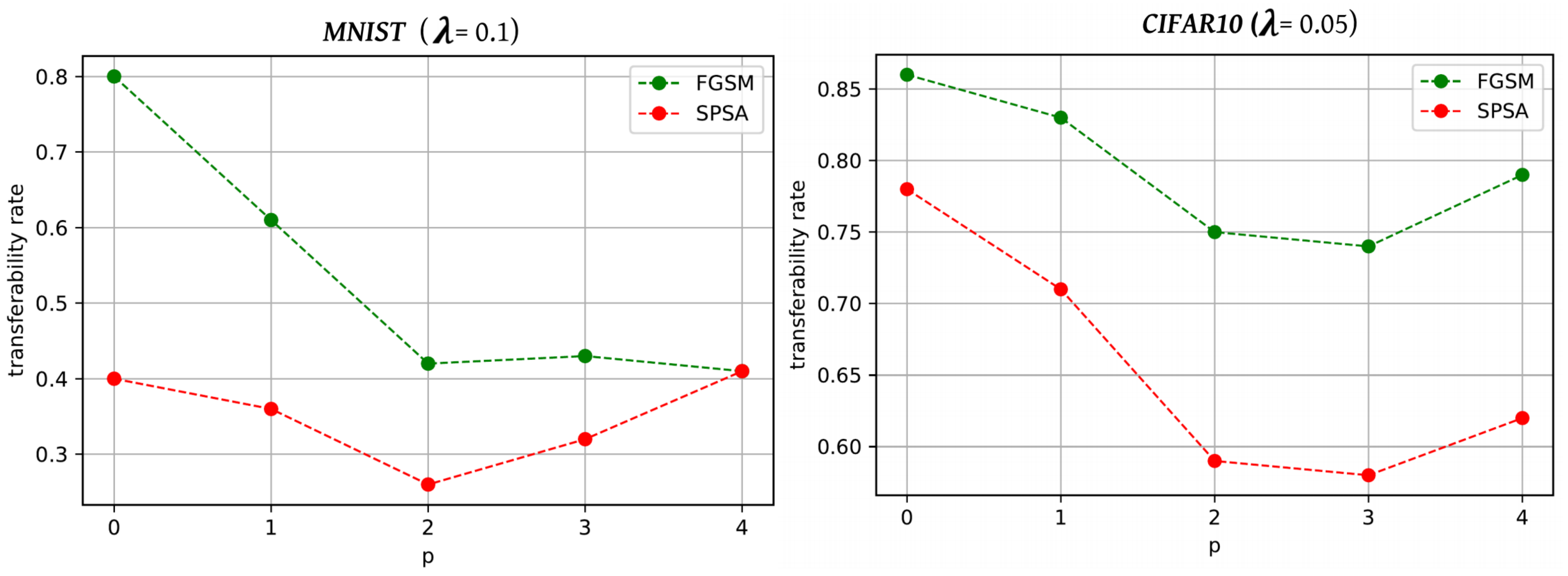}
 
    \caption{\# adversarially trained models $p$ vs. average transferability rate.}
    \label{fig:p-vs-transf}

\end{figure*}

\textbf{Robustness across datasets:}
\sysname{} performs much better on MNIST than CIFAR10. This observation is explained by various factors. First, CNN models are highly accurate on MNIST ($>99\%$) than CIFAR10 ($\approx 84\%$), which makes CIFAR10-CNN more likely to result in misclassifications. Second, across all attacks, on average, \textit{adversarial training} is more effective on MNIST; it not only leads to higher robustness (i.e., $\approx +31\%$ for MNIST compared to $\approx +19\%$ for CIFAR10) it also sacrifices less accuracy on clean MNIST data (i.e.,  $-2.55\%$ for MNIST compared to $-8.26\%$ for CIFAR10). Consequently, the $p$ adversarially-trained student models on MNIST are more robust and accurate than the ones created for CIFAR10. Finally, CIFAR10 adversarial examples are more transferable across student models than MNIST. More results about the transferability factor are detailed in Section \ref{eval:components} with respect to Figures \ref{fig:p-vs-transf} and \ref{fig:lambda-vs-transf}.

\noindent \fbox{\parbox{.96\linewidth}{
 \textbf{Observation 1:} Compared to state-of-the-art empirical defenses such as adversarial training, \sysname{} improves  robustness to adversarial examples on two benchmark datasets (MNIST, CIFAR10) for both white-box and black-box attacks. It does so without sacrificing accuracy on clean data. It is particularly noteworthy that \sysname{} is significantly robust against the C\&W~\cite{CW} attack, one of the strongest white-box attacks on fixed models nowadays.}}

\begin{figure}[t!]
    
    \centering
    \includegraphics[scale=.62]{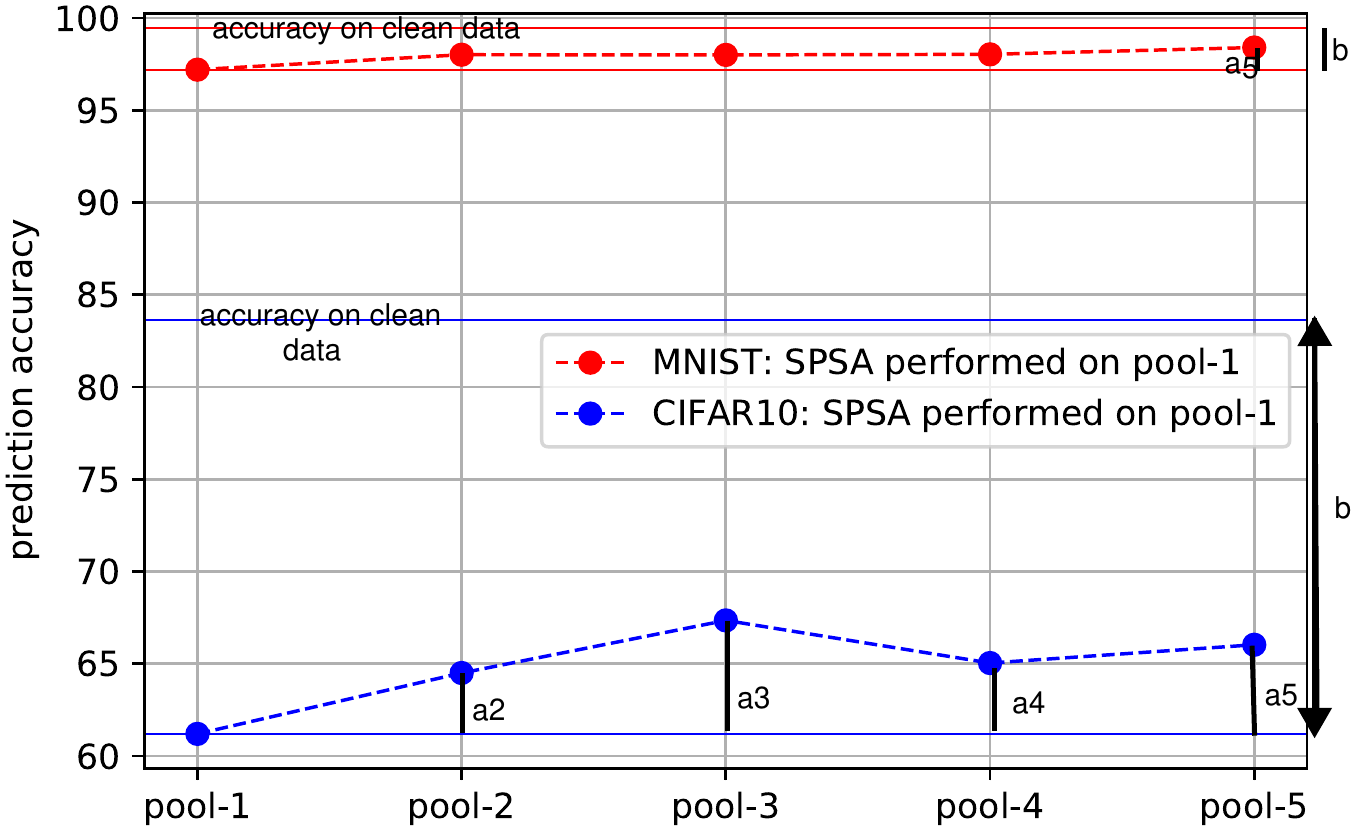}
   \vspace*{0.3em}
    \caption{Impact of model pool renewal on repeating previously successful queries: The prediction accuracy of ulterior pool (i.e., pools 2,3,4, and 5) of models on adversarial examples is generated through multiple SPSA queries on pool-1.}
    \label{fig:repeated}
    \vspace*{-1em}

\end{figure}

\vspace{-1em}

\begin{figure*}[t!]
    
    \centering   
    \includegraphics[width=.97\textwidth]{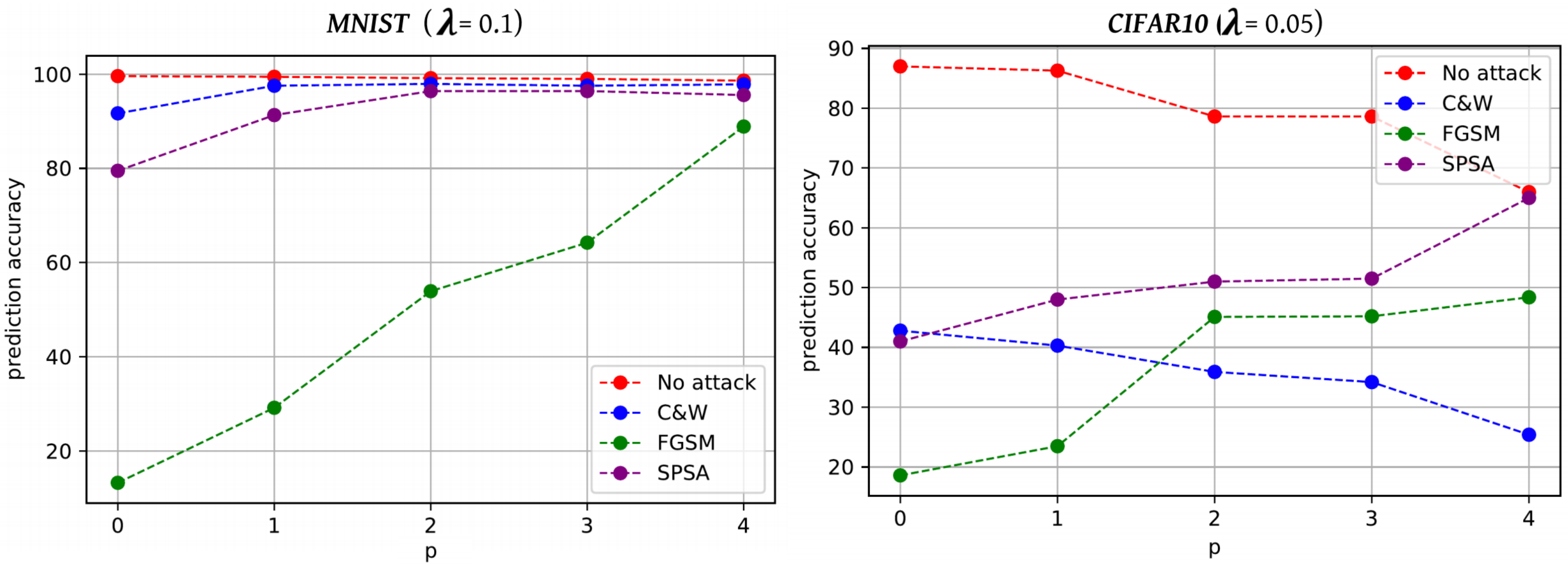}
 
    \caption{\# adversarially trained models $p$ vs. accuracy.}
    \label{fig:p-vs-acc}

\end{figure*}

\subsection{Model Pool Renewal vs. Repeated Attacks}
\label{mtd_impact}
To diagnose the impact of the model pool renewal on the effectiveness of repeated adversarial queries, we perform the SPSA attack by querying only pool-1 of \sysname{}. Then we test the generated adversarial examples on the ulterior pools of models (i.e., pool-2, pool-3, pool-4, and pool-5). For this experiment, we adopt the notation \textit{Failed Repeated Queries (FRQ)} that represents the number of ineffective repeated adversarial queries ($``a"$ in Figure \ref{fig:repeated}) from the total number of  repeated previous adversarial queries ($``b"$ in Figure \ref{fig:repeated}). Results are shown in Figure \ref{fig:repeated}. With respect to the baseline evasion results (i.e., accuracy of pool-1 on SPSA), we observe for both datasets an increase of the accuracy (hence robustness) of ulterior pools (i.e., pool-2,3,4 and 5) on SPSA data generated by querying pool-1. These findings indicate that some of the adversarial examples that were successful on pool-1 are not successful on ulterior pools (i.e., \textit{FRQ}$>0$). More precisely, on average across different pools, $ \approx 87\%$ of the previously effective adversarial examples failed to fool ulterior pools on MNIST ($FRQ = \frac{a}{b} \approx 87\%$). As for CIFAR10, $\approx 21\%$  of previously effective adversarial examples are not successful on ulterior pools ($FRQ = \frac{a}{b} \approx 21\%$). These results reveal the impact of the \textit{model pool renewal} on defending against repeated adversarial queries. However, we note that unlike MNIST, repeated CIFAR10 adversarial queries are more likely to continue to be effective on ulterior pools ($\approx 79\%$ are still effective), which indicates the high transferability rate of SPSA examples across different pools.

\noindent \fbox{\parbox{.96\linewidth}{\textbf{Observation 2:} \sysname{} significantly limits the success of \textit{repeating previously successful adversarial examples} due to the \textit{model pool renewal} step that regularly changes the model pool and creates uncertainty in the eyes of the adversary.}}

At this point, our findings validate the claimed contributions of our approach. Nevertheless, results on CIFAR10 suggest that the evasion transferability of adversarial examples (e.g., FGSM and SPSA) across student models, different pools, and the base model is still a challenge that can hold back \sysname{}'s performance (especially on CIFAR10). These observations invite further investigation into the effectiveness of the measures taken by \sysname{} to reduce transferability rate. Next, we examine the impact of each component in the three steps of \textit{student model pool generation}.

\subsection{Impact of Model Generation Components}
\label{eval:components}
We now focus on the impact of each component of the student model generation steps on \sysname{}'s \textit{robustness against evasion attacks} and \textit{transferability rate across student models}. To that end, we generate different pools of $n=4$ student models using different $0<\lambda<\lambda_{max}$  values and $0\leq p \leq n$ values. In particular, we monitor changes in \textit{prediction accuracy} and the \textit{average transferability rate} across the $n$ student models for different values of $\lambda$ until the maximum bound $\lambda_{max}$ is reached. We recall that $\lambda_{max}$ is reached if the weights perturbation step breaks the student model and leads to the failure of the accuracy recovery step (i.e., failure of Step-2). Additionally, we perform a similar experiment where we try all different possible values of $0\leq p \leq n$. Finally, we evaluate the effectiveness of training each student model on a distinct set and its impact on the reduction of the average transferability rate across models, compared to the case where the same training set $X_{train}$ is utilized for all student models.

\textbf{Retraining on adversarial data.}
For this experiment, we fix a $\lambda$ value that offers acceptable model accuracy and try all possible values of $p$ (i.e., $0\leq p \leq 4$).

\textit{Impact on robustness against adversarial examples:}
Figure \ref{fig:p-vs-acc} shows prediction accuracy of \sysname{} with respect to $0\leq p \leq 4$, for both datasets. For all attacks on MNIST, a $0\rightarrow n$ increase in $p$ leads to an increased prediction accuracy on adversarial data (i.e., more robustness). Similar results are observed for SPSA and FGSM on CIFAR10. However, the prediction accuracy on C\&W data is lower when $p$ is higher which is inline with CIFAR10 results in Table \ref{tab:evasion_res} ($p=8$ vs. $p=9$). As stated earlier (Section \ref{eval:robust}), adversarial training on CIFAR10 leads to a comparatively higher accuracy loss on clean test data. Consequently, the prediction accuracy on clean data decreases when we increase $p$ (``No attack" in Figure \ref{fig:p-vs-acc}). We conclude that, retraining student models on adversarial data is a crucial step to improve the robustness of \sysname{}. However, $p$ needs to be carefully picked to reduce accuracy distortion on clean data caused by adversarial training (especially for CIFAR10), while maximizing \sysname{}'s robustness. From Figure \ref{fig:p-vs-acc}, for both datasets, $p=3$ serves as a practical threshold for $n=4$ (it balances the trade-off between reducing accuracy loss on clean data and increasing prediction accuracy on adversarial data). 

It is noteworthy that for $p=0$, even though no student model is adversarially trained, we observe an increase in \sysname{}'s robustness. For instance, compared to the robustness of the undefended model on MNIST reported in Table \ref{tab:evasion_res}, despite $p=0$ (Figure \ref{fig:p-vs-acc}), the prediction accuracy using a pool of $4$ models is improved on FGSM data ($9.98\% \rightarrow \approx 18\%$). Similar results are observed for C\&W on both datasets (MNIST: $0\% \rightarrow \approx 91\%$, CIFAR10: $0\% \rightarrow \approx 42\%$). These findings, once again, indicate the impact of the MTD aspect on increasing \sysname{}'s robustness against evasion attacks.

\textit{Impact on the evasion transferability across student models:}
Adversarially training a subset of student models serves as an additional defense measure to improve the robustness of \sysname{}. However, it also leads to more diverse student models with respect to the ones that are only trained on clean data which might have an impact on the \textit{average transferability rate} across models. In Figure \ref{fig:p-vs-transf}, we compute the average transferability rate of SPSA and FGSM adversarial examples across student models for each configuration of $0\leq p \leq 4$. We choose SPSA and FGSM for this experiment in view of their high transferability across ML models. Figure \ref{fig:p-vs-transf} shows that, for both datasets, the transferability of both attacks is at its highest rate when $p=0$. Then, it decreases for higher values of $p$ (i.e., $p=1\rightarrow 3$), until it reaches a minimum (i.e., $p=2$ for MNIST and $p=3$ for CIFAR10). Finally, the transferability of both attacks increases again for $p=n=4$. In this final case, all student models are adversarially trained, therefore they are less diverse compared to $p \in \{1,2,3\}$. We conclude that the configuration of the hyper-parameter $p$ has an impact, not only on the overall performance of \sysname{}, but also on the average transferability rate across student models. It is noteworthy that, inline with our previous interpretations in Section \ref{eval:robust}, the transferability rate across student models on CIFAR10 is more important than its value of MNIST, which further explains, \sysname{}'s less performance on CIFAR10.

\begin{figure*}[t!]
    
    \centering   
    \includegraphics[width=.97\textwidth]{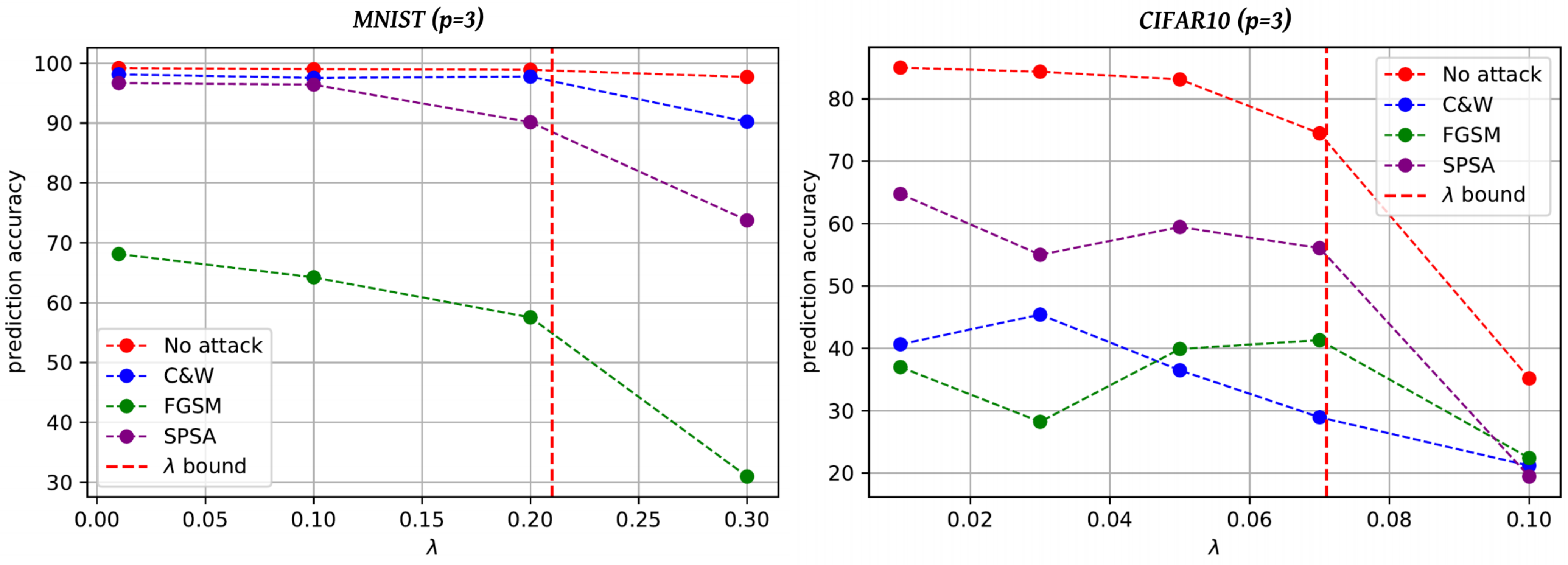}
 
    \caption{Noise scale $\lambda$ vs. accuracy.}
    \label{fig:lambda-vs-acc}

\end{figure*}

\begin{figure*}[t!]
    
    \centering   
    \includegraphics[width=.97\textwidth]{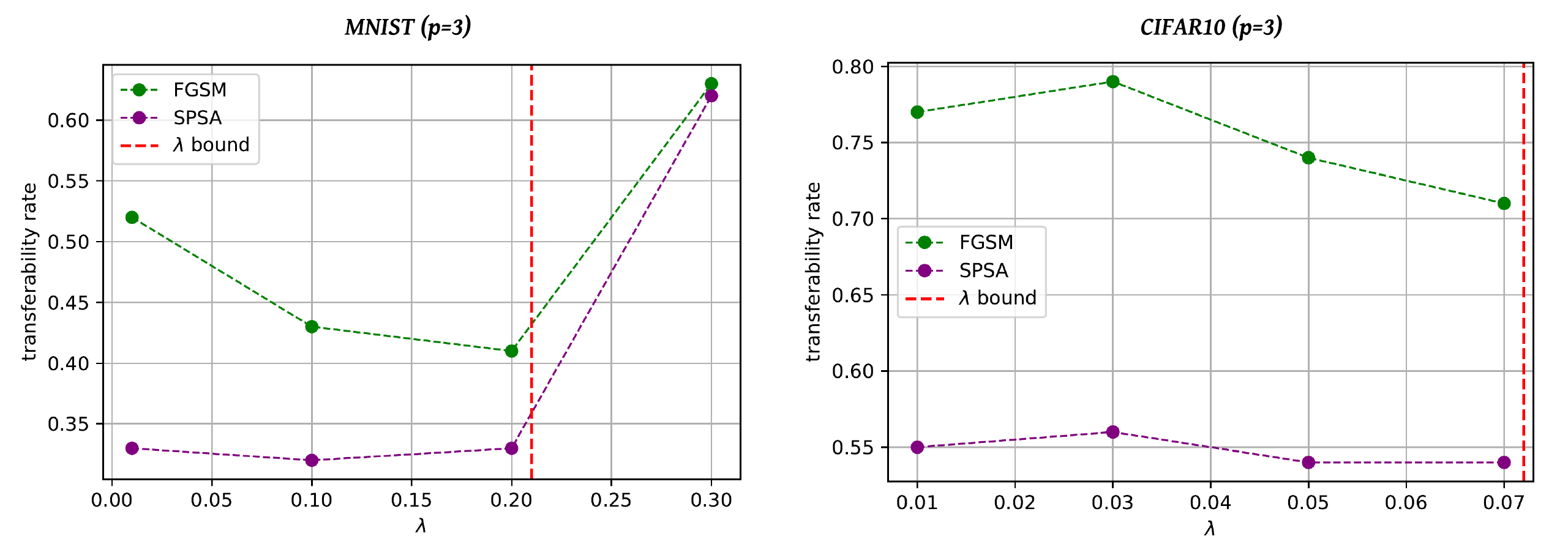}
 
    \caption{Noise scale $\lambda$ vs. average transferability rate.}
    \label{fig:lambda-vs-transf}

\end{figure*}

\textbf{Noise Scale $\lambda$.}
Now we fix $p=3$ (based on the previous results) and we incrementally try different configurations of $\lambda>0$ until we reach a maximum bound $\lambda_{max}$. In Figures \ref{fig:lambda-vs-acc} and \ref{fig:lambda-vs-transf}, with respect to different values of $0< \lambda< \lambda_{max}$, we investigate the impact of the {\em weights perturbation} step on prediction accuracy and average transferability rate across student models. For both datasets, $\lambda_{max}$ is presented as a vertical bound (i.e., red vertical line).

\textit{Impact on transferability rate:} Figure \ref{fig:lambda-vs-transf} indicates that an increase in the noise scale $\lambda$, generally, leads to the decrease of the transferability rate of adversarial examples across student models, for both datasets. This observation is inline with our intuition (in Section \ref{sec: approach}) that higher distortions on the base model weights lead to the generation of more diverse student models.

\textit{Impact on the prediction accuracy:} For MNIST dataset, we observe that higher model weights distortion (i.e., higher $\lambda$) leads to less performance on clean data and on all the studied adversarial data (e.g., C\&W, FGSM and SPSA). As for CIFAR10, Figure \ref{fig:lambda-vs-acc} indicates different results on SPSA and FGSM. For instance, unlike C\&W examples, which are much less transferable, the prediction accuracy on FGSM data reaches its highest when $\lambda=0.05$. Similarly, we observe an increase of the prediction accuracy on SPSA data for $\lambda=0.05$. Next, we further discuss the difference in prediction accuracy patterns on FGSM and SPSA between MNIST and CIFAR10. 

\textit{Impact of evasion transferability across student models on \sysname{} robustness on CIFAR10:}  For $\lambda=0.03$, the prediction accuracy on FGSM and SPSA examples generated on CIFAR10 reaches its minimum (Figure \ref{fig:lambda-vs-acc}), while the transferability rates of both attacks reach their maximum (Figure \ref{fig:lambda-vs-transf}). Furthermore, with respect to all data points of FGSM and SPSA, we notice that the prediction accuracy on FGSM and SPSA increases when the transferability rate across student models decreases. Consequently, for CIFAR10, the transferability rate across models has a crucial impact on the overall performance of \sysname{}. It is noteworthy that, with respect to the different configurations of $p$, Figures \ref{fig:p-vs-acc} and \ref{fig:p-vs-transf}  also indicate the same results of the impact of transferability rate of FGSM and SPSA on their performance on \sysname{}.

\textit{Impact of evasion transferability across student models on \sysname{} robustness on MNIST:} Unlike CIFAR10 dataset, the high transferability rate marked by FGSM and SPSA attacks does not have the same impact on the prediction accuracy. More precisely, the prediction accuracy curves of FGSM and SPSA on MNIST (Figure \ref{fig:lambda-vs-acc}) are different from their counterparts in the transferability rate chart (Figure \ref{fig:lambda-vs-transf}). Furthermore, FGSM and SPSA transferability rate on MNIST is much less than its rate on CIFAR10, which explains the difference of its impact on \sysname{}'s performance.

\textit{Best $\lambda$ configuration:} Our choice of $\lambda=0.1$ represents a tradeoff that balances the reduction of the tansferability rate and the reduction of the accuracy loss, which are conflicting in the case of MNIST. As for CIFAR10, we choose $\lambda=0.05$ to balance between \sysname{} performance against non-transferable attacks (e.g., C\&W) and transferable attacks (e.g., FGSM and SPSA).

\noindent \fbox{\parbox{.96\linewidth}{\textbf{Observation 3:} \sysname{}'s performance is influenced by the values of its hyper-parameters (e.g., $\lambda$, $n$ and $p$). Empirically estimating the optimal configuration of \sysname{} contributes to the reduction of the average transferability rate across student models, which leads to an increased robustness against adversarial examples.}}

\begin{table}[t!]

\centering
  \scalebox{.98}{
   \begin{tabular}{|l!{\color{black}\vrule}l!{\color{black}\vrule}l!{\color{black}\vrule}!{\color{black}\vrule}!{\color{black}\vrule}l!{\color{black}\vrule}l!{\color{black}\vrule}l!{\color{black}\vrule}l!{\color{black}\vrule}}
       \hline
       
&  \multicolumn{2}{c!{\color{black}\vrule}}{\textbf{MNIST-CNN}}
&  \multicolumn{2}{c!{\color{black}\vrule}}{\textbf{CIFAR10-CNN}}\\
       \hline
       & \textbf{FGSM} &  \textbf{SPSA} & \textbf{FGSM} &  \textbf{SPSA}\\ 
       \hline
       Retraining on $X_{train}$ & 0.88 & 0.52&0.95&0.84\\ 
       \hline\hline
       Retraining on $T_i(X_{train})$  &\textbf{0.80} &\textbf{0.40}&\textbf{0.86}&\textbf{0.78}\\ 
       \hline
      
   \end{tabular}}
 
 \vspace*{1em}
\caption{Comparison of average transferability rates of FGSM and SPSA when student models are retrained on $X_{train}$ against considering $T_i(X_{train})$ for retraining ($p=0$).}
\label{tab:transf}
\end{table} 
\textbf{Retraining on transformed data:} As explained in Section \ref{sec: approach}, after Step-1, we perform an \textit{accuracy recovery} step that includes retraining the newly generated student models. We now evaluate to what extent using data transformation reduces the transferability rate of adversarial examples. To that end, we compute the \textit{average transferability rate} of FGSM and SPSA, and we compare it with the baseline case where all student models are retrained on the same training set $X_{train}$. To precisely diagnose the impact of data transformations on transferability, we exclude the effect of Step-3 by using $p=0$, in addition to the same $\lambda$ configuration adopted before. Therefore, we note that the following results do not represent the actual transferability rates of \sysname{} student models (covered in previous discussions). Results reported in Table \ref{tab:transf} show that performing different data transformations $T_i$ on the training set $X_{train}$ before retraining leads to more diverse student models. For instance, we observe $\approx -8,5\%$ less transferable FGSM examples on average across both datasets and an average of $\approx -9\%$ less transferable SPSA examples. However, such reduction is not sufficient to completely bypass the transferability challenge. This is especially observed on CIFAR10. 

\noindent \fbox{\parbox{.96\linewidth}{\textbf{Observation 4: } \textit{Training data transformation} is effective at reducing \textit{average transferability rate}.}}

However, our results on CIFAR10 reveal that \sysname{} might struggle against highly transferable attacks (e.g., FGSM) which indicates that the transferability challenge has room for improvement. Prior work examined the adversarial transferability phenomenon \cite{transferability16, WhyTransfer19}. Yet, rigorous theoretical analysis or explanation for transferability is still lacking in the literature which makes it more difficult to limit transferability of adversarial examples.

\section{Conclusion}\label{sec: concl}
While prior defenses against adversarial examples aim to defend a fixed target model, \sysname{} takes a moving target defense strategy that sufficiently randomizes information an adversary needs to succeed at fooling ML models with adversarial examples. In \sysname{}, model weights perturbation, data transformation, adversarial training, and dynamic model pool scheduling, and seamless model pool renewal work in tandem to defend adversarial example attacks. 
Our extensive evaluations across white-box and black-box attacks on benchmark datasets suggest \sysname{} significantly outperforms adversarial training in improving robustness of ML models against adversarial examples. It does so while maintaining (at times improving) accuracy on clean data and reducing attack transferability among models in the \sysname{} pool. By tracking the success/failure of repeated attacks across batches of model pools, we further validate the effectiveness of the core MTD strategy in \sysname{} in thwarting repeated/correlated adversarial example attacks. 
Looking forward, we see great potential for moving target strategies as effective countermeasures to thwart attacks against machine learning models. 

\section*{Acknowledgements}
We are grateful to the anonymous reviewers for their insightful feedback that improved this paper.
\bibliographystyle{ACM-Reference-Format}
\bibliography{main}


\begin{thebibliography}{56}


\ifx \showCODEN    \undefined \def \showCODEN     #1{\unskip}     \fi
\ifx \showDOI      \undefined \def \showDOI       #1{#1}\fi
\ifx \showISBNx    \undefined \def \showISBNx     #1{\unskip}     \fi
\ifx \showISBNxiii \undefined \def \showISBNxiii  #1{\unskip}     \fi
\ifx \showISSN     \undefined \def \showISSN      #1{\unskip}     \fi
\ifx \showLCCN     \undefined \def \showLCCN      #1{\unskip}     \fi
\ifx \shownote     \undefined \def \shownote      #1{#1}          \fi
\ifx \showarticletitle \undefined \def \showarticletitle #1{#1}   \fi
\ifx \showURL      \undefined \def \showURL       {\relax}        \fi
\providecommand\bibfield[2]{#2}
\providecommand\bibinfo[2]{#2}
\providecommand\natexlab[1]{#1}
\providecommand\showeprint[2][]{arXiv:#2}

\bibitem[\protect\citeauthoryear{Ali and Eshete}{Ali and Eshete}{2020}]%
        {AdvApprox20}
\bibfield{author}{\bibinfo{person}{Abdullah Ali} {and} \bibinfo{person}{Birhanu
  Eshete}.} \bibinfo{year}{2020}\natexlab{}.
\newblock \showarticletitle{Best-Effort Adversarial Approximation of Black-Box
  Malware Classifiers}. In \bibinfo{booktitle}{\emph{Security and Privacy in
  Communication Networks - 16th {EAI} International Conference, SecureComm
  2020}}, Vol.~\bibinfo{volume}{335}. \bibinfo{publisher}{Springer},
  \bibinfo{pages}{318--338}.
\newblock


\bibitem[\protect\citeauthoryear{Athalye, Carlini, and Wagner}{Athalye
  et~al\mbox{.}}{2018}]%
        {Gradient-Masking18}
\bibfield{author}{\bibinfo{person}{Anish Athalye}, \bibinfo{person}{Nicholas
  Carlini}, {and} \bibinfo{person}{David~A. Wagner}.}
  \bibinfo{year}{2018}\natexlab{}.
\newblock \showarticletitle{Obfuscated Gradients Give a False Sense of
  Security: Circumventing Defenses to Adversarial Examples}. In
  \bibinfo{booktitle}{\emph{Proceedings of the 35th International Conference on
  Machine Learning, {ICML} 2018}} \emph{(\bibinfo{series}{Proceedings of
  Machine Learning Research}, Vol.~\bibinfo{volume}{80})}.
  \bibinfo{publisher}{{PMLR}}, \bibinfo{pages}{274--283}.
\newblock


\bibitem[\protect\citeauthoryear{Bhagoji, Cullina, Sitawarin, and
  Mittal}{Bhagoji et~al\mbox{.}}{2018}]%
        {Compression18}
\bibfield{author}{\bibinfo{person}{Arjun~Nitin Bhagoji},
  \bibinfo{person}{Daniel Cullina}, \bibinfo{person}{Chawin Sitawarin}, {and}
  \bibinfo{person}{Prateek Mittal}.} \bibinfo{year}{2018}\natexlab{}.
\newblock \showarticletitle{Enhancing robustness of machine learning systems
  via data transformations}. In \bibinfo{booktitle}{\emph{52nd Annual
  Conference on Information Sciences and Systems, {CISS} 2018}}.
  \bibinfo{publisher}{{IEEE}}, \bibinfo{pages}{1--5}.
\newblock


\bibitem[\protect\citeauthoryear{Biggio, Corona, Maiorca, Nelson, Srndic,
  Laskov, Giacinto, and Roli}{Biggio et~al\mbox{.}}{2013}]%
        {Biggio-ECML13}
\bibfield{author}{\bibinfo{person}{Battista Biggio}, \bibinfo{person}{Igino
  Corona}, \bibinfo{person}{Davide Maiorca}, \bibinfo{person}{Blaine Nelson},
  \bibinfo{person}{Nedim Srndic}, \bibinfo{person}{Pavel Laskov},
  \bibinfo{person}{Giorgio Giacinto}, {and} \bibinfo{person}{Fabio Roli}.}
  \bibinfo{year}{2013}\natexlab{}.
\newblock \showarticletitle{Evasion Attacks against Machine Learning at Test
  Time}. In \bibinfo{booktitle}{\emph{Machine Learning and Knowledge Discovery
  in Databases - European Conference, {ECML} {PKDD} 2013}}.
  \bibinfo{pages}{387--402}.
\newblock


\bibitem[\protect\citeauthoryear{Buckman, Roy, Raffel, and Goodfellow}{Buckman
  et~al\mbox{.}}{2018}]%
        {Thermo-Encode18}
\bibfield{author}{\bibinfo{person}{Jacob Buckman}, \bibinfo{person}{Aurko Roy},
  \bibinfo{person}{Colin Raffel}, {and} \bibinfo{person}{Ian~J. Goodfellow}.}
  \bibinfo{year}{2018}\natexlab{}.
\newblock \showarticletitle{Thermometer Encoding: One Hot Way To Resist
  Adversarial Examples}. In \bibinfo{booktitle}{\emph{6th International
  Conference on Learning Representations, {ICLR} 2018}}.
  \bibinfo{publisher}{OpenReview.net}.
\newblock


\bibitem[\protect\citeauthoryear{Carlini, Athalye, Papernot, Brendel, Rauber,
  Tsipras, Goodfellow, Madry, and Kurakin}{Carlini et~al\mbox{.}}{2019}]%
        {carlini2019evaluating}
\bibfield{author}{\bibinfo{person}{Nicholas Carlini}, \bibinfo{person}{Anish
  Athalye}, \bibinfo{person}{Nicolas Papernot}, \bibinfo{person}{Wieland
  Brendel}, \bibinfo{person}{Jonas Rauber}, \bibinfo{person}{Dimitris Tsipras},
  \bibinfo{person}{Ian Goodfellow}, \bibinfo{person}{Aleksander Madry}, {and}
  \bibinfo{person}{Alexey Kurakin}.} \bibinfo{year}{2019}\natexlab{}.
\newblock \bibinfo{title}{On Evaluating Adversarial Robustness}.
\newblock
\newblock
\showeprint[arxiv]{1902.06705}~[cs.LG]


\bibitem[\protect\citeauthoryear{Carlini and Wagner}{Carlini and
  Wagner}{2017a}]%
        {Carlini-Breaking17}
\bibfield{author}{\bibinfo{person}{Nicholas Carlini} {and}
  \bibinfo{person}{David~A. Wagner}.} \bibinfo{year}{2017}\natexlab{a}.
\newblock \showarticletitle{Adversarial Examples Are Not Easily Detected:
  Bypassing Ten Detection Methods}. In \bibinfo{booktitle}{\emph{Proceedings of
  the 10th {ACM} Workshop on Artificial Intelligence and Security, AISec@CCS
  2017}}. \bibinfo{publisher}{{ACM}}, \bibinfo{pages}{3--14}.
\newblock


\bibitem[\protect\citeauthoryear{Carlini and Wagner}{Carlini and
  Wagner}{2017b}]%
        {CW}
\bibfield{author}{\bibinfo{person}{Nicholas Carlini} {and}
  \bibinfo{person}{David~A. Wagner}.} \bibinfo{year}{2017}\natexlab{b}.
\newblock \showarticletitle{Towards Evaluating the Robustness of Neural
  Networks}. In \bibinfo{booktitle}{\emph{2017 {IEEE} Symposium on Security and
  Privacy, {SP} 2017}}. \bibinfo{pages}{39--57}.
\newblock


\bibitem[\protect\citeauthoryear{Chen, Jordan, and Wainwright}{Chen
  et~al\mbox{.}}{2020}]%
        {HSJA20}
\bibfield{author}{\bibinfo{person}{Jianbo Chen}, \bibinfo{person}{Michael~I.
  Jordan}, {and} \bibinfo{person}{Martin~J. Wainwright}.}
  \bibinfo{year}{2020}\natexlab{}.
\newblock \showarticletitle{HopSkipJumpAttack: {A} Query-Efficient
  Decision-Based Attack}. In \bibinfo{booktitle}{\emph{2020 {IEEE} Symposium on
  Security and Privacy, {SP} 2020}}. \bibinfo{publisher}{{IEEE}},
  \bibinfo{pages}{1277--1294}.
\newblock


\bibitem[\protect\citeauthoryear{Cohen, Rosenfeld, and Kolter}{Cohen
  et~al\mbox{.}}{2019}]%
        {RandomSmoothing19}
\bibfield{author}{\bibinfo{person}{Jeremy~M. Cohen}, \bibinfo{person}{Elan
  Rosenfeld}, {and} \bibinfo{person}{J.~Zico Kolter}.}
  \bibinfo{year}{2019}\natexlab{}.
\newblock \showarticletitle{Certified Adversarial Robustness via Randomized
  Smoothing}. In \bibinfo{booktitle}{\emph{Proceedings of the 36th
  International Conference on Machine Learning, {ICML} 2019}}
  \emph{(\bibinfo{series}{Proceedings of Machine Learning Research},
  Vol.~\bibinfo{volume}{97})}. \bibinfo{publisher}{{PMLR}},
  \bibinfo{pages}{1310--1320}.
\newblock


\bibitem[\protect\citeauthoryear{da~Silva, Berriel, Badue, de~Souza, and
  Oliveira{-}Santos}{da~Silva et~al\mbox{.}}{2018}]%
        {CopyCat18}
\bibfield{author}{\bibinfo{person}{Jacson Rodrigues~Correia da Silva},
  \bibinfo{person}{Rodrigo~Ferreira Berriel}, \bibinfo{person}{Claudine Badue},
  \bibinfo{person}{Alberto~Ferreira de Souza}, {and} \bibinfo{person}{Thiago
  Oliveira{-}Santos}.} \bibinfo{year}{2018}\natexlab{}.
\newblock \showarticletitle{Copycat {CNN:} Stealing Knowledge by Persuading
  Confession with Random Non-Labeled Data}. In \bibinfo{booktitle}{\emph{2018
  International Joint Conference on Neural Networks, {IJCNN} 2018}}.
  \bibinfo{publisher}{{IEEE}}, \bibinfo{pages}{1--8}.
\newblock


\bibitem[\protect\citeauthoryear{Dahl, Yu, Deng, and Acero}{Dahl
  et~al\mbox{.}}{2012}]%
        {DL-Speech2012}
\bibfield{author}{\bibinfo{person}{G.~E. Dahl}, \bibinfo{person}{Dong Yu},
  \bibinfo{person}{Li Deng}, {and} \bibinfo{person}{A. Acero}.}
  \bibinfo{year}{2012}\natexlab{}.
\newblock \showarticletitle{Context-Dependent Pre-Trained Deep Neural Networks
  for Large-Vocabulary Speech Recognition}.
\newblock \bibinfo{journal}{\emph{{IEEE} Transactions on Audio, Speech, and
  Language Processing}} \bibinfo{volume}{20}, \bibinfo{number}{1}
  (\bibinfo{date}{Jan.} \bibinfo{year}{2012}), \bibinfo{pages}{30--42}.
\newblock


\bibitem[\protect\citeauthoryear{Das, Shanbhogue, Chen, Hohman, Chen, Kounavis,
  and Chau}{Das et~al\mbox{.}}{2017}]%
        {Compression17}
\bibfield{author}{\bibinfo{person}{Nilaksh Das}, \bibinfo{person}{Madhuri
  Shanbhogue}, \bibinfo{person}{Shang{-}Tse Chen}, \bibinfo{person}{Fred
  Hohman}, \bibinfo{person}{Li Chen}, \bibinfo{person}{Michael~E. Kounavis},
  {and} \bibinfo{person}{Duen~Horng Chau}.} \bibinfo{year}{2017}\natexlab{}.
\newblock \showarticletitle{Keeping the Bad Guys Out: Protecting and
  Vaccinating Deep Learning with {JPEG} Compression}.
\newblock \bibinfo{journal}{\emph{CoRR}}  \bibinfo{volume}{abs/1705.02900}
  (\bibinfo{year}{2017}).
\newblock


\bibitem[\protect\citeauthoryear{Demontis, Melis, Pintor, Jagielski, Biggio,
  Oprea, Nita{-}Rotaru, and Roli}{Demontis et~al\mbox{.}}{2019}]%
        {WhyTransfer19}
\bibfield{author}{\bibinfo{person}{Ambra Demontis}, \bibinfo{person}{Marco
  Melis}, \bibinfo{person}{Maura Pintor}, \bibinfo{person}{Matthew Jagielski},
  \bibinfo{person}{Battista Biggio}, \bibinfo{person}{Alina Oprea},
  \bibinfo{person}{Cristina Nita{-}Rotaru}, {and} \bibinfo{person}{Fabio
  Roli}.} \bibinfo{year}{2019}\natexlab{}.
\newblock \showarticletitle{{Why Do Adversarial Attacks Transfer? Explaining
  Transferability of Evasion and Poisoning Attacks}}. In
  \bibinfo{booktitle}{\emph{28th {USENIX} Security Symposium, {USENIX} Security
  2019}}. \bibinfo{publisher}{{USENIX} Association}, \bibinfo{pages}{321--338}.
\newblock


\bibitem[\protect\citeauthoryear{Dong, Liao, Pang, Su, Zhu, Hu, and Li}{Dong
  et~al\mbox{.}}{2018}]%
        {MIM}
\bibfield{author}{\bibinfo{person}{Yinpeng Dong}, \bibinfo{person}{Fangzhou
  Liao}, \bibinfo{person}{Tianyu Pang}, \bibinfo{person}{Hang Su},
  \bibinfo{person}{Jun Zhu}, \bibinfo{person}{Xiaolin Hu}, {and}
  \bibinfo{person}{Jianguo Li}.} \bibinfo{year}{2018}\natexlab{}.
\newblock \showarticletitle{Boosting Adversarial Attacks With Momentum}. In
  \bibinfo{booktitle}{\emph{2018 {IEEE} Conference on Computer Vision and
  Pattern Recognition, {CVPR} 2018, Salt Lake City, UT, USA, June 18-22,
  2018}}. \bibinfo{pages}{9185--9193}.
\newblock


\bibitem[\protect\citeauthoryear{Evtimov, Eykholt, Fernandes, Kohno, Li,
  Prakash, Rahmati, and Song}{Evtimov et~al\mbox{.}}{2017}]%
        {AV-Physical-Attack17}
\bibfield{author}{\bibinfo{person}{Ivan Evtimov}, \bibinfo{person}{Kevin
  Eykholt}, \bibinfo{person}{Earlence Fernandes}, \bibinfo{person}{Tadayoshi
  Kohno}, \bibinfo{person}{Bo Li}, \bibinfo{person}{Atul Prakash},
  \bibinfo{person}{Amir Rahmati}, {and} \bibinfo{person}{Dawn Song}.}
  \bibinfo{year}{2017}\natexlab{}.
\newblock \showarticletitle{Robust Physical-World Attacks on Machine Learning
  Models}.
\newblock \bibinfo{journal}{\emph{CoRR}}  \bibinfo{volume}{abs/1707.08945}
  (\bibinfo{year}{2017}).
\newblock


\bibitem[\protect\citeauthoryear{Gallagher, Biernacki, Chen, Aweke, Yitbarek,
  Aga, Harris, Xu, Kasikci, Bertacco, Malik, Tiwari, and Austin}{Gallagher
  et~al\mbox{.}}{2019}]%
        {Morpheus19}
\bibfield{author}{\bibinfo{person}{Mark Gallagher}, \bibinfo{person}{Lauren
  Biernacki}, \bibinfo{person}{Shibo Chen}, \bibinfo{person}{Zelalem~Birhanu
  Aweke}, \bibinfo{person}{Salessawi~Ferede Yitbarek},
  \bibinfo{person}{Misiker~Tadesse Aga}, \bibinfo{person}{Austin Harris},
  \bibinfo{person}{Zhixing Xu}, \bibinfo{person}{Baris Kasikci},
  \bibinfo{person}{Valeria Bertacco}, \bibinfo{person}{Sharad Malik},
  \bibinfo{person}{Mohit Tiwari}, {and} \bibinfo{person}{Todd~M. Austin}.}
  \bibinfo{year}{2019}\natexlab{}.
\newblock \showarticletitle{Morpheus: {A} Vulnerability-Tolerant Secure
  Architecture Based on Ensembles of Moving Target Defenses with Churn}. In
  \bibinfo{booktitle}{\emph{Proceedings of the Twenty-Fourth International
  Conference on Architectural Support for Programming Languages and Operating
  Systems, {ASPLOS} 2019, Providence, RI, USA, April 13-17, 2019}}.
  \bibinfo{publisher}{{ACM}}, \bibinfo{pages}{469--484}.
\newblock


\bibitem[\protect\citeauthoryear{Gao, Wang, Tan, Zhu, Zhang, Fessler,
  Vermeulen, and Wang}{Gao et~al\mbox{.}}{2019}]%
        {DeepCC2019}
\bibfield{author}{\bibinfo{person}{Feng Gao}, \bibinfo{person}{Wei Wang},
  \bibinfo{person}{Miaomiao Tan}, \bibinfo{person}{Lina Zhu},
  \bibinfo{person}{Yuchen Zhang}, \bibinfo{person}{Evelyn Fessler},
  \bibinfo{person}{Louis Vermeulen}, {and} \bibinfo{person}{Xin Wang}.}
  \bibinfo{year}{2019}\natexlab{}.
\newblock \showarticletitle{{DeepCC}: a novel deep learning-based framework for
  cancer molecular subtype classification}.
\newblock \bibinfo{journal}{\emph{Oncogenesis}} \bibinfo{volume}{8},
  \bibinfo{number}{9} (\bibinfo{date}{Aug.} \bibinfo{year}{2019}).
\newblock
\urldef\tempurl%
\url{https://doi.org/10.1038/s41389-019-0157-8}
\showDOI{\tempurl}


\bibitem[\protect\citeauthoryear{Goodfellow}{Goodfellow}{2019}]%
        {goodfellow2019research}
\bibfield{author}{\bibinfo{person}{Ian Goodfellow}.}
  \bibinfo{year}{2019}\natexlab{}.
\newblock \bibinfo{title}{A Research Agenda: Dynamic Models to Defend Against
  Correlated Attacks}.
\newblock
\newblock
\showeprint[arxiv]{1903.06293}~[cs.LG]


\bibitem[\protect\citeauthoryear{Goodfellow, Shlens, and Szegedy}{Goodfellow
  et~al\mbox{.}}{2015}]%
        {FGSM}
\bibfield{author}{\bibinfo{person}{Ian~J. Goodfellow},
  \bibinfo{person}{Jonathon Shlens}, {and} \bibinfo{person}{Christian
  Szegedy}.} \bibinfo{year}{2015}\natexlab{}.
\newblock \showarticletitle{Explaining and Harnessing Adversarial Examples}. In
  \bibinfo{booktitle}{\emph{3rd International Conference on Learning
  Representations, {ICLR}}}.
\newblock


\bibitem[\protect\citeauthoryear{Gu and Rigazio}{Gu and Rigazio}{2015}]%
        {Early-Defense14}
\bibfield{author}{\bibinfo{person}{Shixiang Gu} {and} \bibinfo{person}{Luca
  Rigazio}.} \bibinfo{year}{2015}\natexlab{}.
\newblock \showarticletitle{Towards Deep Neural Network Architectures Robust to
  Adversarial Examples}. In \bibinfo{booktitle}{\emph{3rd International
  Conference on Learning Representations, {ICLR} 2015}}.
\newblock


\bibitem[\protect\citeauthoryear{Guo, Rana, Ciss{\'{e}}, and van~der
  Maaten}{Guo et~al\mbox{.}}{2018}]%
        {Rand18}
\bibfield{author}{\bibinfo{person}{Chuan Guo}, \bibinfo{person}{Mayank Rana},
  \bibinfo{person}{Moustapha Ciss{\'{e}}}, {and} \bibinfo{person}{Laurens
  van~der Maaten}.} \bibinfo{year}{2018}\natexlab{}.
\newblock \showarticletitle{Countering Adversarial Images using Input
  Transformations}. In \bibinfo{booktitle}{\emph{6th International Conference
  on Learning Representations, {ICLR} 2018}}.
  \bibinfo{publisher}{OpenReview.net}.
\newblock


\bibitem[\protect\citeauthoryear{He, Wei, Chen, Carlini, and Song}{He
  et~al\mbox{.}}{2017}]%
        {Carlini-BreakingUsenix17}
\bibfield{author}{\bibinfo{person}{Warren He}, \bibinfo{person}{James Wei},
  \bibinfo{person}{Xinyun Chen}, \bibinfo{person}{Nicholas Carlini}, {and}
  \bibinfo{person}{Dawn Song}.} \bibinfo{year}{2017}\natexlab{}.
\newblock \showarticletitle{Adversarial Example Defense: Ensembles of Weak
  Defenses are not Strong}. In \bibinfo{booktitle}{\emph{11th {USENIX} Workshop
  on Offensive Technologies, {WOOT} 2017, Vancouver, BC, Canada, August 14-15,
  2017}}. \bibinfo{publisher}{{USENIX} Association}.
\newblock


\bibitem[\protect\citeauthoryear{Hu and Tan}{Hu and Tan}{2017}]%
        {MalGAN17}
\bibfield{author}{\bibinfo{person}{Weiwei Hu} {and} \bibinfo{person}{Ying
  Tan}.} \bibinfo{year}{2017}\natexlab{}.
\newblock \showarticletitle{Generating Adversarial Malware Examples for
  Black-Box Attacks Based on {GAN}}.
\newblock \bibinfo{journal}{\emph{CoRR}}  \bibinfo{volume}{abs/1702.05983}
  (\bibinfo{year}{2017}).
\newblock


\bibitem[\protect\citeauthoryear{Huang, Xu, Schuurmans, and
  Szepesv{\'{a}}ri}{Huang et~al\mbox{.}}{2015}]%
        {Early-Defense15}
\bibfield{author}{\bibinfo{person}{Ruitong Huang}, \bibinfo{person}{Bing Xu},
  \bibinfo{person}{Dale Schuurmans}, {and} \bibinfo{person}{Csaba
  Szepesv{\'{a}}ri}.} \bibinfo{year}{2015}\natexlab{}.
\newblock \showarticletitle{Learning with a Strong Adversary}.
\newblock \bibinfo{journal}{\emph{CoRR}}  \bibinfo{volume}{abs/1511.03034}
  (\bibinfo{year}{2015}).
\newblock


\bibitem[\protect\citeauthoryear{Jajodia, Ghosh, Swarup, Wang, and
  Wang}{Jajodia et~al\mbox{.}}{2011}]%
        {MTD-book}
\bibfield{editor}{\bibinfo{person}{Sushil Jajodia}, \bibinfo{person}{Anup~K.
  Ghosh}, \bibinfo{person}{Vipin Swarup}, \bibinfo{person}{Cliff Wang}, {and}
  \bibinfo{person}{Xiaoyang~Sean Wang}} (Eds.).
  \bibinfo{year}{2011}\natexlab{}.
\newblock \bibinfo{booktitle}{\emph{Moving Target Defense - Creating Asymmetric
  Uncertainty for Cyber Threats}}. \bibinfo{series}{Advances in Information
  Security}, Vol.~\bibinfo{volume}{54}.
\newblock \bibinfo{publisher}{Springer}.
\newblock


\bibitem[\protect\citeauthoryear{Kolosnjaji, Demontis, Biggio, Maiorca,
  Giacinto, Eckert, and Roli}{Kolosnjaji et~al\mbox{.}}{2018}]%
        {MalConvEvade18}
\bibfield{author}{\bibinfo{person}{Bojan Kolosnjaji}, \bibinfo{person}{Ambra
  Demontis}, \bibinfo{person}{Battista Biggio}, \bibinfo{person}{Davide
  Maiorca}, \bibinfo{person}{Giorgio Giacinto}, \bibinfo{person}{Claudia
  Eckert}, {and} \bibinfo{person}{Fabio Roli}.}
  \bibinfo{year}{2018}\natexlab{}.
\newblock \showarticletitle{Adversarial Malware Binaries: Evading Deep Learning
  for Malware Detection in Executables}. In \bibinfo{booktitle}{\emph{26th
  European Signal Processing Conference, {EUSIPCO}}}.
  \bibinfo{pages}{533--537}.
\newblock


\bibitem[\protect\citeauthoryear{Krizhevsky, Nair, and Hinton}{Krizhevsky
  et~al\mbox{.}}{[n.d.]}]%
        {cifar}
\bibfield{author}{\bibinfo{person}{Alex Krizhevsky}, \bibinfo{person}{Vinod
  Nair}, {and} \bibinfo{person}{Geoffrey Hinton}.}
  \bibinfo{year}{[n.d.]}\natexlab{}.
\newblock \showarticletitle{CIFAR-10 (Canadian Institute for Advanced
  Research)}.
\newblock  (\bibinfo{year}{[n.\,d.]}).
\newblock
\urldef\tempurl%
\url{http://www.cs.toronto.edu/~kriz/cifar.html}
\showURL{%
\tempurl}


\bibitem[\protect\citeauthoryear{Krizhevsky, Sutskever, and Hinton}{Krizhevsky
  et~al\mbox{.}}{2017}]%
        {ImageNet}
\bibfield{author}{\bibinfo{person}{Alex Krizhevsky}, \bibinfo{person}{Ilya
  Sutskever}, {and} \bibinfo{person}{Geoffrey~E. Hinton}.}
  \bibinfo{year}{2017}\natexlab{}.
\newblock \showarticletitle{ImageNet classification with deep convolutional
  neural networks}.
\newblock \bibinfo{journal}{\emph{Commun. {ACM}}} \bibinfo{volume}{60},
  \bibinfo{number}{6} (\bibinfo{year}{2017}), \bibinfo{pages}{84--90}.
\newblock


\bibitem[\protect\citeauthoryear{Kurakin, Goodfellow, and Bengio}{Kurakin
  et~al\mbox{.}}{2017}]%
        {kurakin2017adversarial}
\bibfield{author}{\bibinfo{person}{Alexey Kurakin}, \bibinfo{person}{Ian
  Goodfellow}, {and} \bibinfo{person}{Samy Bengio}.}
  \bibinfo{year}{2017}\natexlab{}.
\newblock \bibinfo{title}{Adversarial Machine Learning at Scale}.
\newblock
\newblock
\showeprint[arxiv]{1611.01236}~[cs.CV]


\bibitem[\protect\citeauthoryear{Kurakin, Goodfellow, and Bengio}{Kurakin
  et~al\mbox{.}}{2016}]%
        {BIM}
\bibfield{author}{\bibinfo{person}{Alexey Kurakin}, \bibinfo{person}{Ian~J.
  Goodfellow}, {and} \bibinfo{person}{Samy Bengio}.}
  \bibinfo{year}{2016}\natexlab{}.
\newblock \showarticletitle{Adversarial Machine Learning at Scale}.
\newblock \bibinfo{journal}{\emph{CoRR}}  \bibinfo{volume}{abs/1611.01236}
  (\bibinfo{year}{2016}).
\newblock


\bibitem[\protect\citeauthoryear{LeCun, Cortes, and Burges}{LeCun
  et~al\mbox{.}}{2020}]%
        {MNIST}
\bibfield{author}{\bibinfo{person}{Yan LeCun}, \bibinfo{person}{Corinna
  Cortes}, {and} \bibinfo{person}{Christopher~J.C. Burges}.}
  \bibinfo{year}{2020}\natexlab{}.
\newblock \bibinfo{title}{The MNIST Database of Handwritten Digits}.
\newblock \bibinfo{howpublished}{\url{http://yann.lecun.com/exdb/mnist/}}.
\newblock


\bibitem[\protect\citeauthoryear{Lecuyer, Atlidakis, Geambasu, Hsu, and
  Jana}{Lecuyer et~al\mbox{.}}{2019}]%
        {lecuyer2019certified}
\bibfield{author}{\bibinfo{person}{Mathias Lecuyer}, \bibinfo{person}{Vaggelis
  Atlidakis}, \bibinfo{person}{Roxana Geambasu}, \bibinfo{person}{Daniel Hsu},
  {and} \bibinfo{person}{Suman Jana}.} \bibinfo{year}{2019}\natexlab{}.
\newblock \bibinfo{title}{Certified Robustness to Adversarial Examples with
  Differential Privacy}.
\newblock
\newblock
\showeprint[arxiv]{1802.03471}~[stat.ML]


\bibitem[\protect\citeauthoryear{Lei, Zhang, Tan, Zhang, and Liu}{Lei
  et~al\mbox{.}}{[n.d.]}]%
        {MTD-Survey18}
\bibfield{author}{\bibinfo{person}{Cheng Lei}, \bibinfo{person}{Hongqi Zhang},
  \bibinfo{person}{Jing{-}Lei Tan}, \bibinfo{person}{Yu{-}Chen Zhang}, {and}
  \bibinfo{person}{Xiao{-}Hu Liu}.} \bibinfo{year}{[n.d.]}\natexlab{}.
\newblock \showarticletitle{Moving Target Defense Techniques: {A} Survey}.
\newblock \bibinfo{journal}{\emph{Secur. Commun. Networks}}
  \bibinfo{volume}{2018} (\bibinfo{year}{[n.\,d.]}).
\newblock


\bibitem[\protect\citeauthoryear{Li, Chen, Wang, and Carin}{Li
  et~al\mbox{.}}{2019}]%
        {Certified-AdditiveNoise19}
\bibfield{author}{\bibinfo{person}{Bai Li}, \bibinfo{person}{Changyou Chen},
  \bibinfo{person}{Wenlin Wang}, {and} \bibinfo{person}{Lawrence Carin}.}
  \bibinfo{year}{2019}\natexlab{}.
\newblock \showarticletitle{Certified Adversarial Robustness with Additive
  Noise}. In \bibinfo{booktitle}{\emph{Advances in Neural Information
  Processing Systems 32: Annual Conference on Neural Information Processing
  Systems 2019, NeurIPS 2019}}. \bibinfo{pages}{9459--9469}.
\newblock


\bibitem[\protect\citeauthoryear{Luo, Boix, Roig, Poggio, and Zhao}{Luo
  et~al\mbox{.}}{2015}]%
        {Rand15}
\bibfield{author}{\bibinfo{person}{Yan Luo}, \bibinfo{person}{Xavier Boix},
  \bibinfo{person}{Gemma Roig}, \bibinfo{person}{Tomaso~A. Poggio}, {and}
  \bibinfo{person}{Qi Zhao}.} \bibinfo{year}{2015}\natexlab{}.
\newblock \showarticletitle{Foveation-based Mechanisms Alleviate Adversarial
  Examples}.
\newblock \bibinfo{journal}{\emph{CoRR}}  \bibinfo{volume}{abs/1511.06292}
  (\bibinfo{year}{2015}).
\newblock


\bibitem[\protect\citeauthoryear{Madry, Makelov, Schmidt, Tsipras, and
  Vladu}{Madry et~al\mbox{.}}{2017}]%
        {PGSM}
\bibfield{author}{\bibinfo{person}{Aleksander Madry},
  \bibinfo{person}{Aleksandar Makelov}, \bibinfo{person}{Ludwig Schmidt},
  \bibinfo{person}{Dimitris Tsipras}, {and} \bibinfo{person}{Adrian Vladu}.}
  \bibinfo{year}{2017}\natexlab{}.
\newblock \showarticletitle{Towards Deep Learning Models Resistant to
  Adversarial Attacks}.
\newblock \bibinfo{journal}{\emph{CoRR}}  \bibinfo{volume}{abs/1706.06083}
  (\bibinfo{year}{2017}).
\newblock


\bibitem[\protect\citeauthoryear{Orekondy, Schiele, and Fritz}{Orekondy
  et~al\mbox{.}}{2018}]%
        {orekondy2018knockoff}
\bibfield{author}{\bibinfo{person}{Tribhuvanesh Orekondy},
  \bibinfo{person}{Bernt Schiele}, {and} \bibinfo{person}{Mario Fritz}.}
  \bibinfo{year}{2018}\natexlab{}.
\newblock \bibinfo{title}{Knockoff Nets: Stealing Functionality of Black-Box
  Models}.
\newblock
\newblock
\showeprint[arxiv]{1812.02766}~[cs.CV]


\bibitem[\protect\citeauthoryear{Papernot, McDaniel, Wu, Jha, and
  Swami}{Papernot et~al\mbox{.}}{2016c}]%
        {distillation}
\bibfield{author}{\bibinfo{person}{Nicolas Papernot}, \bibinfo{person}{Patrick
  McDaniel}, \bibinfo{person}{Xi Wu}, \bibinfo{person}{Somesh Jha}, {and}
  \bibinfo{person}{Ananthram Swami}.} \bibinfo{year}{2016}\natexlab{c}.
\newblock \showarticletitle{Distillation as a Defense to Adversarial
  Perturbations Against Deep Neural Networks}. \bibinfo{pages}{582--597}.
\newblock
\urldef\tempurl%
\url{https://doi.org/10.1109/SP.2016.41}
\showDOI{\tempurl}


\bibitem[\protect\citeauthoryear{Papernot, McDaniel, and Goodfellow}{Papernot
  et~al\mbox{.}}{2016a}]%
        {transferability16}
\bibfield{author}{\bibinfo{person}{Nicolas Papernot},
  \bibinfo{person}{Patrick~D. McDaniel}, {and} \bibinfo{person}{Ian~J.
  Goodfellow}.} \bibinfo{year}{2016}\natexlab{a}.
\newblock \showarticletitle{Transferability in Machine Learning: from Phenomena
  to Black-Box Attacks using Adversarial Samples}.
\newblock \bibinfo{journal}{\emph{CoRR}}  \bibinfo{volume}{abs/1605.07277}
  (\bibinfo{year}{2016}).
\newblock


\bibitem[\protect\citeauthoryear{Papernot, McDaniel, Goodfellow, Jha, Celik,
  and Swami}{Papernot et~al\mbox{.}}{2016b}]%
        {Practical-black-box16}
\bibfield{author}{\bibinfo{person}{Nicolas Papernot},
  \bibinfo{person}{Patrick~D. McDaniel}, \bibinfo{person}{Ian~J. Goodfellow},
  \bibinfo{person}{Somesh Jha}, \bibinfo{person}{Z.~Berkay Celik}, {and}
  \bibinfo{person}{Ananthram Swami}.} \bibinfo{year}{2016}\natexlab{b}.
\newblock \showarticletitle{{Practical Black-Box Attacks against Deep Learning
  Systems using Adversarial Examples}}.
\newblock \bibinfo{journal}{\emph{CoRR}}  \bibinfo{volume}{abs/1602.02697}
  (\bibinfo{year}{2016}).
\newblock


\bibitem[\protect\citeauthoryear{Qian, Shao, Wang, Lin, Guo, Gu, Wang, and
  Wu}{Qian et~al\mbox{.}}{2020}]%
        {EI-MTD}
\bibfield{author}{\bibinfo{person}{Yaguan Qian}, \bibinfo{person}{Qiqi Shao},
  \bibinfo{person}{Jiamin Wang}, \bibinfo{person}{Xiang Lin},
  \bibinfo{person}{Yankai Guo}, \bibinfo{person}{Zhaoquan Gu},
  \bibinfo{person}{Bin Wang}, {and} \bibinfo{person}{Chunming Wu}.}
  \bibinfo{year}{2020}\natexlab{}.
\newblock \showarticletitle{{EI-MTD:} Moving Target Defense for Edge
  Intelligence against Adversarial Attacks}.
\newblock \bibinfo{journal}{\emph{CoRR}}  \bibinfo{volume}{abs/2009.10537}
  (\bibinfo{year}{2020}).
\newblock


\bibitem[\protect\citeauthoryear{Raff, Barker, Sylvester, Brandon, Catanzaro,
  and Nicholas}{Raff et~al\mbox{.}}{2018}]%
        {malconv18}
\bibfield{author}{\bibinfo{person}{Edward Raff}, \bibinfo{person}{Jon Barker},
  \bibinfo{person}{Jared Sylvester}, \bibinfo{person}{Robert Brandon},
  \bibinfo{person}{Bryan Catanzaro}, {and} \bibinfo{person}{Charles~K.
  Nicholas}.} \bibinfo{year}{2018}\natexlab{}.
\newblock \showarticletitle{Malware Detection by Eating a Whole {EXE}}. In
  \bibinfo{booktitle}{\emph{The Workshops of the The Thirty-Second {AAAI}
  Conference on Artificial Intelligence}}. \bibinfo{pages}{268--276}.
\newblock


\bibitem[\protect\citeauthoryear{Raghunathan, Steinhardt, and
  Liang}{Raghunathan et~al\mbox{.}}{2020}]%
        {raghunathan2020certified}
\bibfield{author}{\bibinfo{person}{Aditi Raghunathan}, \bibinfo{person}{Jacob
  Steinhardt}, {and} \bibinfo{person}{Percy Liang}.}
  \bibinfo{year}{2020}\natexlab{}.
\newblock \bibinfo{title}{Certified Defenses against Adversarial Examples}.
\newblock
\newblock
\showeprint[arxiv]{1801.09344}~[cs.LG]


\bibitem[\protect\citeauthoryear{Sallab, Abdou, Perot, and Yogamani}{Sallab
  et~al\mbox{.}}{2017}]%
        {DL-autnonmous17}
\bibfield{author}{\bibinfo{person}{Ahmad~El Sallab}, \bibinfo{person}{Mohammed
  Abdou}, \bibinfo{person}{Etienne Perot}, {and} \bibinfo{person}{Senthil~Kumar
  Yogamani}.} \bibinfo{year}{2017}\natexlab{}.
\newblock \showarticletitle{Deep Reinforcement Learning framework for
  Autonomous Driving}.
\newblock \bibinfo{journal}{\emph{CoRR}}  \bibinfo{volume}{abs/1704.02532}
  (\bibinfo{year}{2017}).
\newblock


\bibitem[\protect\citeauthoryear{Samuel~Kotz}{Samuel~Kotz}{2012}]%
        {Laplace}
\bibfield{author}{\bibinfo{person}{Krzystof~Podgorski Samuel~Kotz,
  Tomasz~Kozubowski}.} \bibinfo{year}{2012}\natexlab{}.
\newblock \bibinfo{title}{The Laplace Distribution and Generalizations: A
  Revisit with Applications to Communications, Economics, Engineering, and
  Finance}.
\newblock
\newblock


\bibitem[\protect\citeauthoryear{Sengupta, Chakraborti, and
  Kambhampati}{Sengupta et~al\mbox{.}}{2019}]%
        {MTDeep19}
\bibfield{author}{\bibinfo{person}{Sailik Sengupta}, \bibinfo{person}{Tathagata
  Chakraborti}, {and} \bibinfo{person}{Subbarao Kambhampati}.}
  \bibinfo{year}{2019}\natexlab{}.
\newblock \showarticletitle{MTDeep: Boosting the Security of Deep Neural Nets
  Against Adversarial Attacks with Moving Target Defense}. In
  \bibinfo{booktitle}{\emph{Decision and Game Theory for Security - 10th
  International Conference, GameSec 2019}} \emph{(\bibinfo{series}{Lecture
  Notes in Computer Science}, Vol.~\bibinfo{volume}{11836})}.
  \bibinfo{publisher}{Springer}, \bibinfo{pages}{479--491}.
\newblock


\bibitem[\protect\citeauthoryear{Song, Yan, and Tan}{Song
  et~al\mbox{.}}{2019}]%
        {fMTD19}
\bibfield{author}{\bibinfo{person}{Qun Song}, \bibinfo{person}{Zhenyu Yan},
  {and} \bibinfo{person}{Rui Tan}.} \bibinfo{year}{2019}\natexlab{}.
\newblock \showarticletitle{Moving target defense for embedded deep visual
  sensing against adversarial examples}. In
  \bibinfo{booktitle}{\emph{Proceedings of the 17th Conference on Embedded
  Networked Sensor Systems, SenSys 2019}}. \bibinfo{publisher}{{ACM}},
  \bibinfo{pages}{124--137}.
\newblock


\bibitem[\protect\citeauthoryear{Song, Kim, Nowozin, Ermon, and Kushman}{Song
  et~al\mbox{.}}{2018}]%
        {PixelDefend18}
\bibfield{author}{\bibinfo{person}{Yang Song}, \bibinfo{person}{Taesup Kim},
  \bibinfo{person}{Sebastian Nowozin}, \bibinfo{person}{Stefano Ermon}, {and}
  \bibinfo{person}{Nate Kushman}.} \bibinfo{year}{2018}\natexlab{}.
\newblock \showarticletitle{PixelDefend: Leveraging Generative Models to
  Understand and Defend against Adversarial Examples}. In
  \bibinfo{booktitle}{\emph{6th International Conference on Learning
  Representations, {ICLR} 2018}}.
\newblock


\bibitem[\protect\citeauthoryear{Tian, Yang, and Cai}{Tian
  et~al\mbox{.}}{2018}]%
        {Tian2018DetectingAE}
\bibfield{author}{\bibinfo{person}{Shixin Tian}, \bibinfo{person}{Guolei Yang},
  {and} \bibinfo{person}{Y. Cai}.} \bibinfo{year}{2018}\natexlab{}.
\newblock \showarticletitle{Detecting Adversarial Examples Through Image
  Transformation}. In \bibinfo{booktitle}{\emph{AAAI}}.
\newblock


\bibitem[\protect\citeauthoryear{Tram{\`{e}}r, Kurakin, Papernot, Goodfellow,
  Boneh, and McDaniel}{Tram{\`{e}}r et~al\mbox{.}}{2018}]%
        {EnsembelAdvTrain18}
\bibfield{author}{\bibinfo{person}{Florian Tram{\`{e}}r},
  \bibinfo{person}{Alexey Kurakin}, \bibinfo{person}{Nicolas Papernot},
  \bibinfo{person}{Ian~J. Goodfellow}, \bibinfo{person}{Dan Boneh}, {and}
  \bibinfo{person}{Patrick~D. McDaniel}.} \bibinfo{year}{2018}\natexlab{}.
\newblock \showarticletitle{Ensemble Adversarial Training: Attacks and
  Defenses}. In \bibinfo{booktitle}{\emph{6th International Conference on
  Learning Representations, {ICLR} 2018}}.
\newblock


\bibitem[\protect\citeauthoryear{Tram{\`e}r, Zhang, Juels, Reiter, and
  Ristenpart}{Tram{\`e}r et~al\mbox{.}}{2016}]%
        {Model-Stealing}
\bibfield{author}{\bibinfo{person}{Florian Tram{\`e}r}, \bibinfo{person}{Fan
  Zhang}, \bibinfo{person}{Ari Juels}, \bibinfo{person}{Michael~K. Reiter},
  {and} \bibinfo{person}{Thomas Ristenpart}.} \bibinfo{year}{2016}\natexlab{}.
\newblock \showarticletitle{Stealing Machine Learning Models via Prediction
  APIs}. In \bibinfo{booktitle}{\emph{25th {USENIX} Security Symposium
  ({USENIX} Security 16)}}. \bibinfo{address}{Austin, TX},
  \bibinfo{pages}{601--618}.
\newblock
\showISBNx{978-1-931971-32-4}


\bibitem[\protect\citeauthoryear{Uesato, O'Donoghue, van~den Oord, and
  Kohli}{Uesato et~al\mbox{.}}{2018}]%
        {uesato2018adversarial}
\bibfield{author}{\bibinfo{person}{Jonathan Uesato}, \bibinfo{person}{Brendan
  O'Donoghue}, \bibinfo{person}{Aaron van~den Oord}, {and}
  \bibinfo{person}{Pushmeet Kohli}.} \bibinfo{year}{2018}\natexlab{}.
\newblock \bibinfo{title}{Adversarial Risk and the Dangers of Evaluating
  Against Weak Attacks}.
\newblock
\newblock
\showeprint[arxiv]{1802.05666}~[cs.LG]


\bibitem[\protect\citeauthoryear{Wong and Kolter}{Wong and Kolter}{2018}]%
        {wong2018provable}
\bibfield{author}{\bibinfo{person}{Eric Wong} {and} \bibinfo{person}{J.~Zico
  Kolter}.} \bibinfo{year}{2018}\natexlab{}.
\newblock \bibinfo{title}{Provable defenses against adversarial examples via
  the convex outer adversarial polytope}.
\newblock
\newblock
\showeprint[arxiv]{1711.00851}~[cs.LG]


\bibitem[\protect\citeauthoryear{Xie, Wang, Zhang, Zhou, Xie, and Yuille}{Xie
  et~al\mbox{.}}{2017}]%
        {Cropping17}
\bibfield{author}{\bibinfo{person}{Cihang Xie}, \bibinfo{person}{Jianyu Wang},
  \bibinfo{person}{Zhishuai Zhang}, \bibinfo{person}{Yuyin Zhou},
  \bibinfo{person}{Lingxi Xie}, {and} \bibinfo{person}{Alan~L. Yuille}.}
  \bibinfo{year}{2017}\natexlab{}.
\newblock \showarticletitle{Adversarial Examples for Semantic Segmentation and
  Object Detection}. In \bibinfo{booktitle}{\emph{{IEEE} International
  Conference on Computer Vision, {ICCV} 2017}}. \bibinfo{publisher}{{IEEE}
  Computer Society}, \bibinfo{pages}{1378--1387}.
\newblock


\bibitem[\protect\citeauthoryear{Zantedeschi, Nicolae, and Rawat}{Zantedeschi
  et~al\mbox{.}}{2017}]%
        {Augmentation17}
\bibfield{author}{\bibinfo{person}{Valentina Zantedeschi},
  \bibinfo{person}{Maria{-}Irina Nicolae}, {and} \bibinfo{person}{Ambrish
  Rawat}.} \bibinfo{year}{2017}\natexlab{}.
\newblock \showarticletitle{Efficient Defenses Against Adversarial Attacks}. In
  \bibinfo{booktitle}{\emph{Proceedings of the 10th {ACM} Workshop on
  Artificial Intelligence and Security, AISec@CCS 2017}}.
  \bibinfo{publisher}{{ACM}}, \bibinfo{pages}{39--49}.
\newblock


\end{thebibliography}
\section*{Appendix}
\label{sec: appendix}

\subsection{Base Models Architectures}\label{subsec:base-models}
In Table \ref{tab:models}, we detail the architectures of MNIST-CNN and CIFAR10-CNN.
\begin{table}[h!]

 \begin{center}
    \scalebox{.80}{

   \begin{tabular}{|c|c|} 
   
   \hline
       
      \textbf{Base model} & \textbf{Architecture} \\ \hline
MNIST-CNN & \shortstack{$Conv2d(1, 32, 3, 1)+Relu$;
            $Conv2d(32, 64, 3, 1)+Relu$\\
            $max\_pool2d+dropout(0.25)+flatten$;
            $Linear(9216,128)+Relu$\\
            $dropout(0.5)+Linear(128, 10)$;
            $softmax(1)$}
         \\
         \hline

CIFAR10-CNN& $\shortstack{Conv2d(3,32,3,1)+
            BatchNorm2d(32)+
            ReLu;\\
            Conv2d(32,64, 3, 1)+
            ReLU+
        MaxPool2d\\
        Conv2d(64, 128, 3, 1)+
            BatchNorm2d(128)+
            ReLU\\
            Conv2d(128, 128, 3, 1)+
            ReLU+
            MaxPool2d+
            Dropout2d(p=0.05)\\
            Conv2d(128, 256,3,1)+
            BatchNorm2d(256)+
           ReLU\\
            Conv2d(256, 256, 3, 1)+
            ReLU+
            MaxPool2d\\
            flatten+Dropout(p=0.1)+
            Linear(4096, 1024)+
            ReLU\\
            Linear(1024, 512)+
            ReLU+
            Dropout(p=0.1)+
            Linear(512, 10)
            }$ \\ 
       \hline

   \end{tabular}}
 \end{center}
\vspace{2em}
 \caption{Base models architectures.}\label{tab:models}

\end{table}

\subsection{Attacks}\label{subsec:attacks}

\textbf{Fast-Gradient Sign Method (FGS)}~\cite{FGSM}: This attack is a fast one-step method that crafts an adversarial example. Considering the dot product of the weight vector $\theta$ and an adversarial example (i.e., $x' = x+\delta$), $\theta^\top x' = \theta^\top x + \theta^\top \delta$, the adversarial perturbation causes the activation to grow by $\theta^\top \delta$. Goodfellow et al.~\cite{FGSM} suggested to maximize this increase subject to the maximum perturbation constraint $||\delta||< \epsilon$ by assigning $\delta = sign(\theta)$. Given a sample $x$, the optimal perturbation is given as follows:
\begin{equation}
\label{eq:FGSM}
    \delta^\star = \epsilon . sign(\nabla_xJ(\theta, x, y_{target}))
\end{equation}

\textbf{Carlini-Wagner (C\&W)}~\cite{CW}: This attack is one of the most powerful attacks, where the adversarial example generation problem is formulated as the following optimization problem:
\begin{equation}
\label{eq:BIM}
\begin{array}{rrclcl}
    minimize \quad D(x,x+\delta)\\
    \textrm{s.t.} \quad f(x+\delta) = y_{target}\\
    \quad x+\delta \in [0,1]^n
\end{array}
\end{equation}
The goal is to find a small change $\delta$ such that when added to an image $x$, the image is misclassified (to a targeted class $y_{target}$) by the model but the image is still a valid image. $D$ is some distance metric (e.g $||.||_0$, $||.||_2$ or $||.||_\infty$). Due to the non-linear nature of the classification function $f$, authors defined a simpler objective function $g$ such that $f(x+\delta)=y_{target}$ if and only if $g(x+\delta)<0$. Multiple options of the explicit definition of $g$ are discussed in the paper ~\cite{CW} (e.g., $g(x')=-J(\theta,x',y_{target})+1$). Considering the $p^{th}$ norm as the distance $D$, the optimization problem is simplified as follows:
\begin{equation}
\label{eq:BIM}
\begin{array}{rrclcl}
    minimize \quad ||\delta||_p + c.g(x+\delta)\\
    \textrm{s.t.} \quad x+\delta \in [0,1]^n\\
\end{array}
\end{equation}
where c > 0 is a suitably chosen constant.\\

\textbf{SPSA ~\cite{uesato2018adversarial}:} Simultaneous Perturbation Stochastic Approximation (SPSA) is an attack that re-purposes gradient-free optimization techniques into adversarial example attacks. They explore adversarial risk as a measure of the model’s performance on worst-case inputs. Since the exact adversarial risk is computationally intractable to exactly evaluate, they rather frame commonly used attacks (such as the ones we described earlier) and adversarial evaluation metrics to define a tractable surrogate objective to the true adversarial risk. The details of the algorithm are in~\cite{uesato2018adversarial}.

\textbf{Copycat~\cite{CopyCat18}}: Copycat is a learning-based method to attack a model and steal its knowledge. It considers an adversary that has no access to the training data or classification problem of the target model. However, it uses a surrogate model that has the same architecture of the target model. The process of Copycat is training the surrogate model on a large set of Random Natural Images (ImageNet and Microsoft COCO) labeled (hard-label) by target model. Copycat only uses the hard-labels, it is not necessary to know the probabilities of the target model. In this paper, we use a transformed set of the training data (i.e., MNIST and CIFAR10) of the target model instead of Random Natural Images, which makes the attack stronger and suitable for robustness evaluation in worst case scenarios.

\subsection{Attack Hyper-Parameters}
\label{hyper}
In Table \ref{tab:params}, we specify the parameters used for each studied attack.

\begin{table}[h!]

 \begin{center}
    \scalebox{.97}{

   \begin{tabular}{|c|c|} \toprule

      \textbf{Attack} & \textbf{Hyper-Parameters} \\ \hline
         FGSM &  \shortstack{$lb = 0$, $ub = 1$,\\ which indicate the lower and upper \\ bound of features. These two parameters are\\ specified only for MNIST dataset.}\\
         \hline

       PGD& \shortstack{$lb=0$, $ub=1$, with solver parameters:\\ ($\eta = 0.5$, $\eta\_min=2.0$, $max_{iter}=100$)}. \\
       \hline
      
      SPSA & \shortstack{$learning\_rate=0.01$,   \\$spsa\_samples=128$, $nb_{iter}=10$}. \\
       \hline

   \end{tabular}}
   \vspace{2em}
                 \caption{Attack Hyper-Parameters.}\label{tab:params}

 \end{center}

\end{table} 

\end{document}